%% file: main.tex
\documentclass[10pt]{article}
\usepackage{url}
\usepackage{palatino}
\usepackage{latexsym}
\usepackage{amsmath}
\usepackage{xcolor}
\usepackage{fullpage}
\usepackage{listings}
\usepackage{hyperref}
\usepackage{graphicx}
\usepackage[normalem]{ulem}
\usepackage{wrapfig}
\usepackage{amsfonts,amsthm}
\usepackage[skins]{tcolorbox}
\usepackage{relsize}
\usepackage{enumitem}
\usepackage{titlesec}
\usepackage{xspace}
\usepackage{ifthen}
\usepackage{lipsum}
\usepackage{todonotes}
\usepackage[numbers]{natbib}
\usepackage{adjustbox}

\usepackage[font=small,skip=15pt]{caption} 

\setcitestyle{square}

\usepackage{enumitem}
\setitemize{noitemsep,topsep=0pt,parsep=0pt,partopsep=0pt}
\setenumerate{noitemsep,topsep=0pt,parsep=0pt,partopsep=0pt}

\titlespacing*{\subsection}{0pt}{1pt}{1pt}

\titlespacing{\paragraph}{%
  0pt}{
  0.5\baselineskip}{
  1em}
\titlespacing{\section}{0pt}{2ex}{.5ex}   
\titlespacing{\subsection}{0pt}{0.5ex}{0.5ex}
\titlespacing{\subsubsection}{0pt}{0ex}{0ex}
\titlespacing{\paragraph}{0pt}{0.5ex}{0.5ex}

\definecolor{lightest-gray}{gray}{0.95}
\definecolor{light-gray}{gray}{0.8}
\definecolor{darkgray}{rgb}{0.2, 0.2, 0.2}

\usepackage{listings}
\theoremstyle{definition}

\usepackage{subcaption}

\newcommand{\ignore}[1]{}

\newcommand{\reccox}[2]{\medskip\begin{quote}\noindent {\bf Recommendation #1:~~} {\it #2}\end{quote}\vspace{-1.3ex}}

\begin{document}

\input{titlepage}
\input{executive-summary}  

\newpage
\pagenumbering{arabic}
\setcounter{page}{1}

\input{intro}
\input{human_factors}
\input{robustness-adaptability}
\input{infrastructure}
\input{speech}
\input{lu} 
\input{dialog_interactive} 
\input{synthesis}
\input{policy}

\vspace{11.6cm}
\centerline{\large \bf Acknowledgments}
\smallskip
\noindent
The workshop that led to this report was sponsored by the National Science Foundation, Grant \#IIS1941541. 
We thank Tanya Korelsky for her vision, support, and guidance.  
We also thank the government 
observers who attended the workshop and contributed
to the discussions: Jonathan Fiscus, Susan G. Hill, Nia Peters,
Erion Plaku, Christopher Reardon, and Clare Voss. 
A hearty thanks also to Erin Zaroukian for her careful reading and thoughtful
comments on the writing, Chad M. Smith for illustrating
the title page, Evan Jensen for improving 
the technical figures, and Amber Bennett-Groves for proofreading.

\newpage
\bibliographystyle{plainnat}
\bibliography{references}

\end{document}

%% file: titlepage.tex
\rule{0cm}{1mm}
\vspace{0mm}
\thispagestyle{empty}

\begin{center}{\huge \bf Spoken Language Interaction with Robots}\end{center}

\centerline{\it --- --- ---}

\begin{center}{\huge \bf Research Issues and Recommendations}\end{center}

\vspace{2cm}
\begin{figure}[thb]
    \centering
\includegraphics[width=10cm]{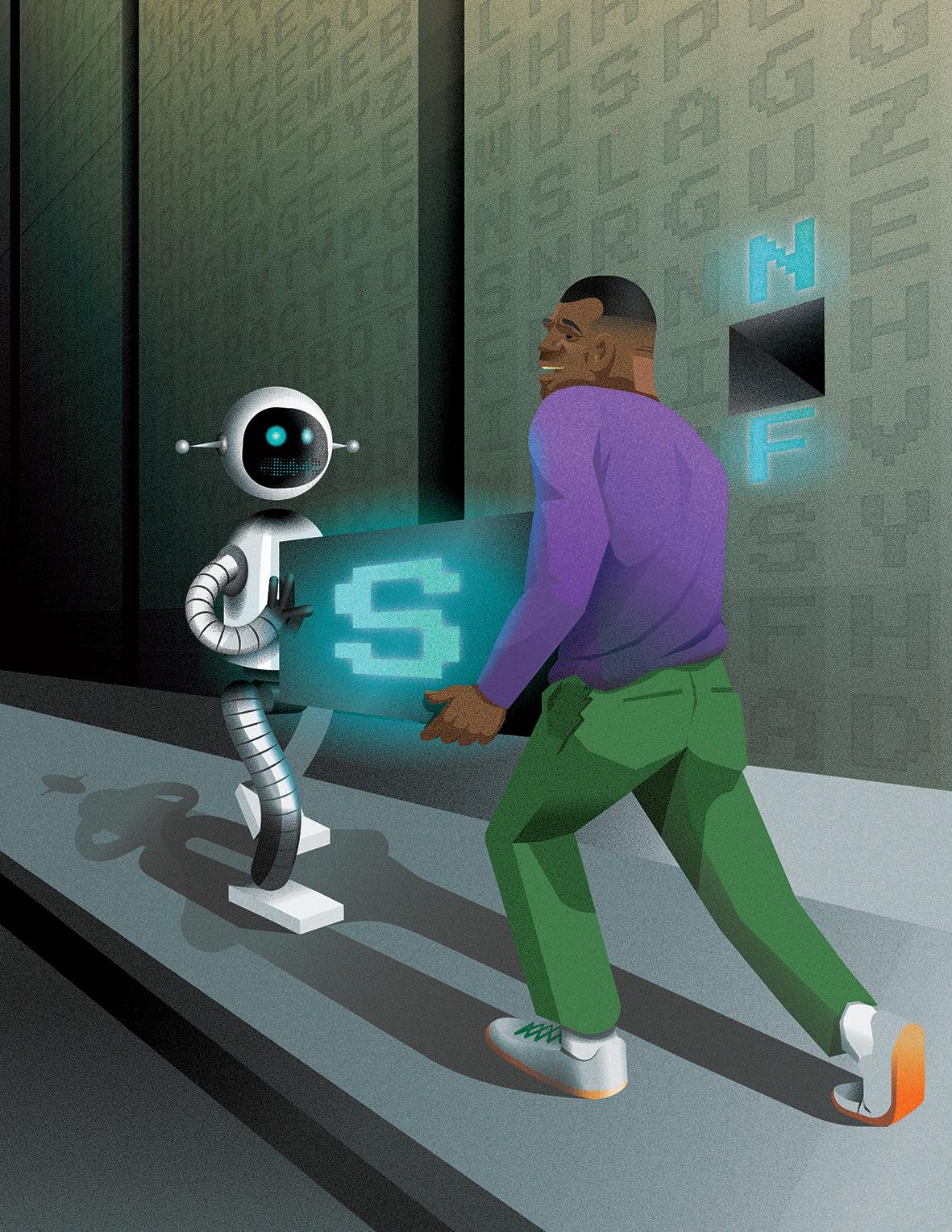}      
    \label{fig:cover-image}
\end{figure}

\vspace{2cm}

\centerline{\Large \bf Matthew Marge, Carol Espy-Wilson, Nigel G. Ward}

\clearpage
\thispagestyle{empty}

\rule{0cm}{1mm}
\vspace{0mm}

\noindent{\Large \bf Spoken Language Interaction with Robots:}    

\medskip
\noindent{\Large \bf Research Issues and Recommendations}

\medskip\noindent
Report from the NSF Future Directions Workshop, October 2019

\vspace{1cm}
\newcommand{\au}[2]{{\noindent \bf #1}, #2\vspace{.65mm}\\}
\noindent\au{Matthew Marge}{DEVCOM Army Research Laboratory} 
\noindent\au{Carol Espy-Wilson}{University of Maryland, College Park}
\noindent\au{Nigel G. Ward}{University of Texas at El Paso}

\vspace{5mm}
\noindent
{\bf Workshop Participants and Report Contributors }

\vspace{.4cm}
\au{Abeer Alwan}{University of California at Los Angeles}
\au{Yoav Artzi}{Cornell University}
\au{Mohit Bansal}{University of North Carolina at Chapel Hill}
\au{Gil Blankenship}{University of Maryland}
\au{Joyce Chai}{University of Michigan}
\au{Hal Daum\'e III}{University of Maryland, College Park}
\au{Debadeepta Dey}{Microsoft}
\au{Mary Harper}{DEVCOM Army Research Laboratory, emerita}
\au{Thomas Howard}{University of Rochester}
\au{Casey Kennington}{Boise State University}
\au{Tatiana (Tanya) Korelsky}{National Science Foundation}
\au{Ivana Kruijff-Korbayov\'a}{DFKI}
\au{Dinesh Manocha}{University of Maryland, College Park}
\au{Cynthia Matuszek}{University of Maryland, Baltimore County}
\au{Ross Mead}{Semio}
\au{Raymond Mooney}{University of Texas at Austin}
\au{Roger K.~Moore}{University of Sheffield}
\au{Mari Ostendorf}{University of Washington}
\au{Heather Pon-Barry}{Mount Holyoke College}
\au{Alexander I.~Rudnicky}{Carnegie Mellon University}
\au{Matthias Scheutz}{Tufts University}
\au{Robert St.\ Amant}{DEVCOM Army Research Laboratory}
\au{Tong Sun}{Adobe}
\au{Stefanie Tellex}{Brown University}
\au{David Traum}{USC Institute for Creative Technologies}
\au{Zhou Yu}{University of California at Davis}

\vspace{8mm}
\noindent\today  

\clearpage 

%% file: executive-summary.tex
\thispagestyle{empty}
\rule{1mm}{0mm}
\vspace{.7cm}

\centerline{\large \bf Executive Summary}
\bigskip 


\noindent
\begin{quote} With robotics rapidly advancing, more effective human-robot
interaction is increasingly needed to realize the full
potential of robots for society.  While spoken language must  be part of the
solution,  our ability to provide spoken language interaction
capabilities is still very limited.  The National Science Foundation accordingly convened a
workshop, bringing together speech, language, and robotics researchers
to discuss what needs to be done.  The result is this report, in which
we identify key scientific and engineering advances needed to enable
effective spoken language interaction with robotics.

\medskip
Our recommendations broadly relate to eight general
themes.  First, meeting human needs requires addressing new challenges in
speech technology and user experience design.  Second, this requires
better models of the social and interactive aspects of language use.
Third, for robustness, robots need higher-bandwidth communication with
users and better handling of uncertainty, including simultaneous
consideration of multiple hypotheses and goals.  Fourth, more powerful
adaptation methods are needed, to enable robots to communicate in new
environments, for new tasks, and with diverse user populations,
without extensive re-engineering or the collection of massive training
data.  Fifth, since robots are embodied, speech should function
together with other communication modalities, such as gaze, gesture,
posture, and motion.  Sixth, since robots operate in complex
environments, speech components need access to rich yet efficient
representations of what the robot knows about objects, locations,
noise sources, the user, and other humans.  Seventh, since robots
operate in real time, their speech and language processing components
must also.  Eighth, in addition to more research, we need more work on
infrastructure and resources, including shareable software modules and
internal interfaces, inexpensive hardware, baseline systems, and
diverse corpora.

\medskip
Research and development that prioritizes these issues will, we
believe, provide a solid foundation for the creation of speech-capable robots that
are easy and effective for humans to work with.
\end{quote}




\vspace{1.cm}
\newcommand{\tocline}[1]{\noindent{\bf #1}\vspace{2mm}}
\begin{quote}
\tocline{1.~ Introduction}

\tocline{2.~ User Experience Design }

\tocline{3.~ Robustness and Adaptability}

\tocline{4.~ Infrastructure}

\tocline{5.~ Audio and Speech Processing, Speech Recognition, and Behavior Signal Analysis}

\tocline{6.~ Language Understanding}

\tocline{7.~ Dialogue and Human Communication Dynamics}

\tocline{8.~ Language Generation and Speech Synthesis}

\tocline{9.~ Policy}
\end{quote}

%% file: intro.tex
\section{Introduction}

As robotics advances, spoken language interaction is becoming increasingly necessary.
Yet robot researchers often find it difficult to incorporate speech processing capabilities,
and speech researchers seldom appreciate the special needs of robot applications.
We hope to help members of both tribes, as well as social scientists, other researchers, and subject matter experts to better understand the difficulties,
possibilities, and research issues in speech for robots; to catalyze new
research projects in this area; and to thereby bring us closer to
the vision of truly satisfying spoken language interaction with robots.

Thus this report aims to identify the key challenges that robotics
brings for spoken interaction, and the key issues in designing robot
systems able to make effective use of speech.  We make 31
recommendations, relating broadly to issues of human needs, sociality
and interaction, robustness, adaptation, multimodality,
representations, timing, and infrastructure.  We address these
recommendations to funding agencies, leaders in industry, principal
investigators, graduate students, developers, and system integrators.

Our work started with a two-day meeting, October 10 and 11, 2019,
hosted by Carol Espy-Wilson at the University of Maryland with support
from Tanya Korelsky's program at the National Science Foundation,
followed by a year of discussion and reflection.  Unlike more
general roadmaps for research in robotics, dialogue systems, 
artificial intelligence, and related areas
\cite{amershi2019guidelines,beckerle2019robotic,nsf-multimedia,robotics-roadmap,eskenazi2020report,ai-roadmap19,sheridan2016,vinciarelli12,ward-devault-aimag,wiltshire2013towards,yang2018grand}, 
this report focuses on issues that are especially critical or
especially challenging for speech in robotics.
This report is intended to complement the work of
the similarly-themed Dagstuhl workshop on Spoken Language Interaction with Virtual Agents and Robots \cite{slivar2020}: their report provides more information on recent research advances and trends, while here we focus on making specific recommendations.

\subsection*{Why Spoken Language Interaction with Robots?}

\begin{figure}[t]
    \centering
\includegraphics[width=4.5in]{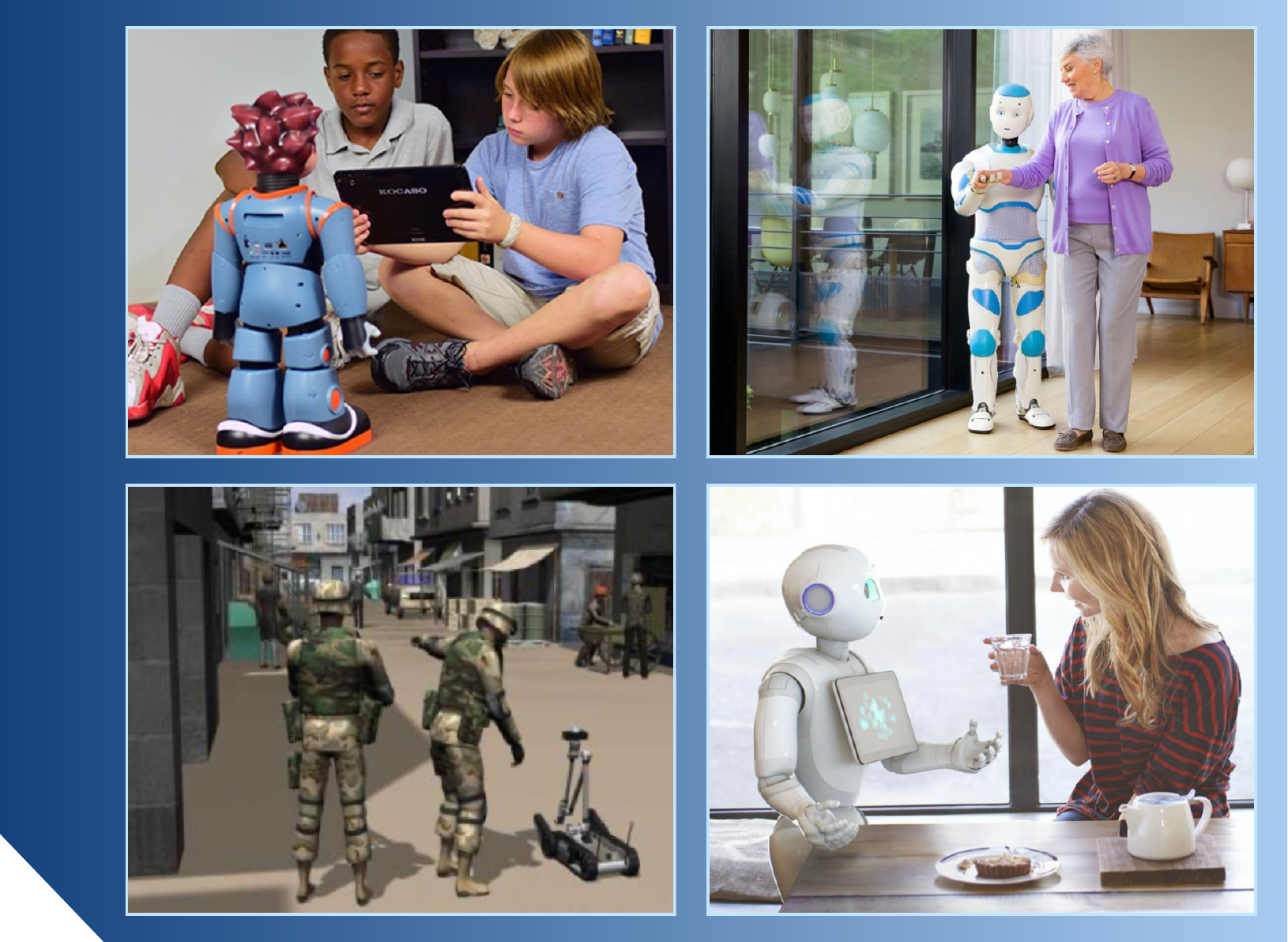}      
\caption{Spoken language interaction with robots across four application areas: education, healthcare, field assistance, and entertainment~\cite{iso-robots-rescue}. Image credits: RoboKind, SoftBank Robotics, and the United States Army. Reproduced with permission.}
    \label{fig:apps}
\end{figure}

\noindent 
Across a wide range of applications, spoken language interaction with robots has great promise. As shown in Figure~\ref{fig:apps}, among the most immediate applications are in education, healthcare, field assistance (both civilian and military), and the consumer market. 
Robot tutors can provide a sense of presence, which can stimulate interests and joy in learning and improve learning outcomes~\cite{bainbridge11}. 
Moreover, the learning experience with a robot can be tailored to individual students to shape the learning process~\cite{ramachandran19}.
Speech-capable robots can amplify a healthcare worker's capabilities in remote care (e.g., a rehabilitation robot tracking and assisting a patient with physical therapy)~\cite{gockley2006encouraging}, as well as create engaging social interactions with elderly patients that may be isolated due to medical reasons~\cite{broekens2009assistive}. 
Robots fielded in search-and-rescue, humanitarian relief, or reconnaissance scenarios also require an expressive interface like spoken language, especially when situations call for rapid specification of robot goals (such as when the human teammate is engaged in other activities)~\cite{kruijff2015tradr}. Speech also allows an operator to oversee and command teams of robots doing tasks without the use of a handheld controller~\cite{marge2019research}.
Finally, consumer robotics (including but not limited to entertainment, security, and household applications) represents the fastest-growing market segment, projected to be a \$23-billion market by 2025~\cite{Wolfgang2017}. Products such as Pepper, Jibo, and Anki Vector have shown the promise of speech interfaces to increase the ease-of-use of robots in homes by non-expert users.

Reasons why spoken language interaction with robots will greatly benefit human society include:
\begin{itemize}
\item Among the various ways to exchange information with robots,
spoken language has the potential to often be the fastest
  and most efficient.  Speed is critical for robots capable of
  interacting with people in real time. Especially in operations where
  time is of the essence, slow performance is equivalent to failure.
  Speed is required not only during the action, but also in the human-robot
  communication, both prior to and during execution.

\item Spoken language interaction will enable new dimensions of
  human-robot cooperative action, such as the realtime coordination of
  physical actions by human and robot.

\item Spoken language interaction is socially potent, and will enable
  robots to engage in more motivating, satisfying, and reassuring
  interactions, for example, when tutoring children, caring for the
  sick, and supporting people in dangerous environments.
  
\item As robots become more capable, people will \emph{expect} speech
  to be the primary way to interact with robots.
  
  \item Robots that you can talk with may be simply better liked, a critical consideration for
  consumer robotics.

\item Robots can be better communicators than disembodied voices \cite{deng2019embodiment};
  being co-present, a robot's gestures and actions can reinforce or
  clarify a message, help manage turn-taking more efficiently, convey
  nuances of stance or intent, and so on.

\item Building speech-capable robots is an intellectual grand
  challenge that will drive advances across the speech and language
  sciences and beyond.
\end{itemize}

Not every robot needs speech, but speech serves functions that are
essential in many scenarios.  Meeting these needs is, however, beyond
the current state of the art.

\subsection*{Why Don't We Have It Yet?}

\noindent At first glance, speech for robots seems like it should be a
simple matter of plugging in some off-the-shelf modules and getting a
talking robot \cite{moore2015talking}.  But it's not that easy.  This
report will discuss the reasons at length, but here 
we give an initial overview of the relevant properties
of robots and spoken communication.

\medskip
\textbf{What is a robot, in essence?}  While in many ways a robot is like any
other AI system that may need to converse with a human, there are some
fundamental differences \cite{nsf-robotics}.  Notably, in general:

\begin{enumerate}
\item A robot is situated; it exists at a specific point in space, and
  interacts with the environment, affecting it and being affected.
\item A robot provides affordances; its physical embodiment affects
  how people perceive its actions, speech, and capabilities, and affects how they
  choose to interact with it.
\item A robot has very limited abilities, in both perception and
  action; it is never able to fully control or fully understand the situation.
\item A robot exists at a specific moment in time, but a time where
  everything may be in a state of change --- the environment, the
  robot's current plans and ongoing actions, what it's hearing, what
  it's saying, and so on.
\end{enumerate}

Not every robot brings unique challenges for speech --- a robot that
just sits on a desk, chatting and smiling, can work much like any
other conversational agent --- but as robots become more capable, speech
becomes more challenging.  

\begin{figure}[t]
    \centering
\includegraphics[width=4.5in]{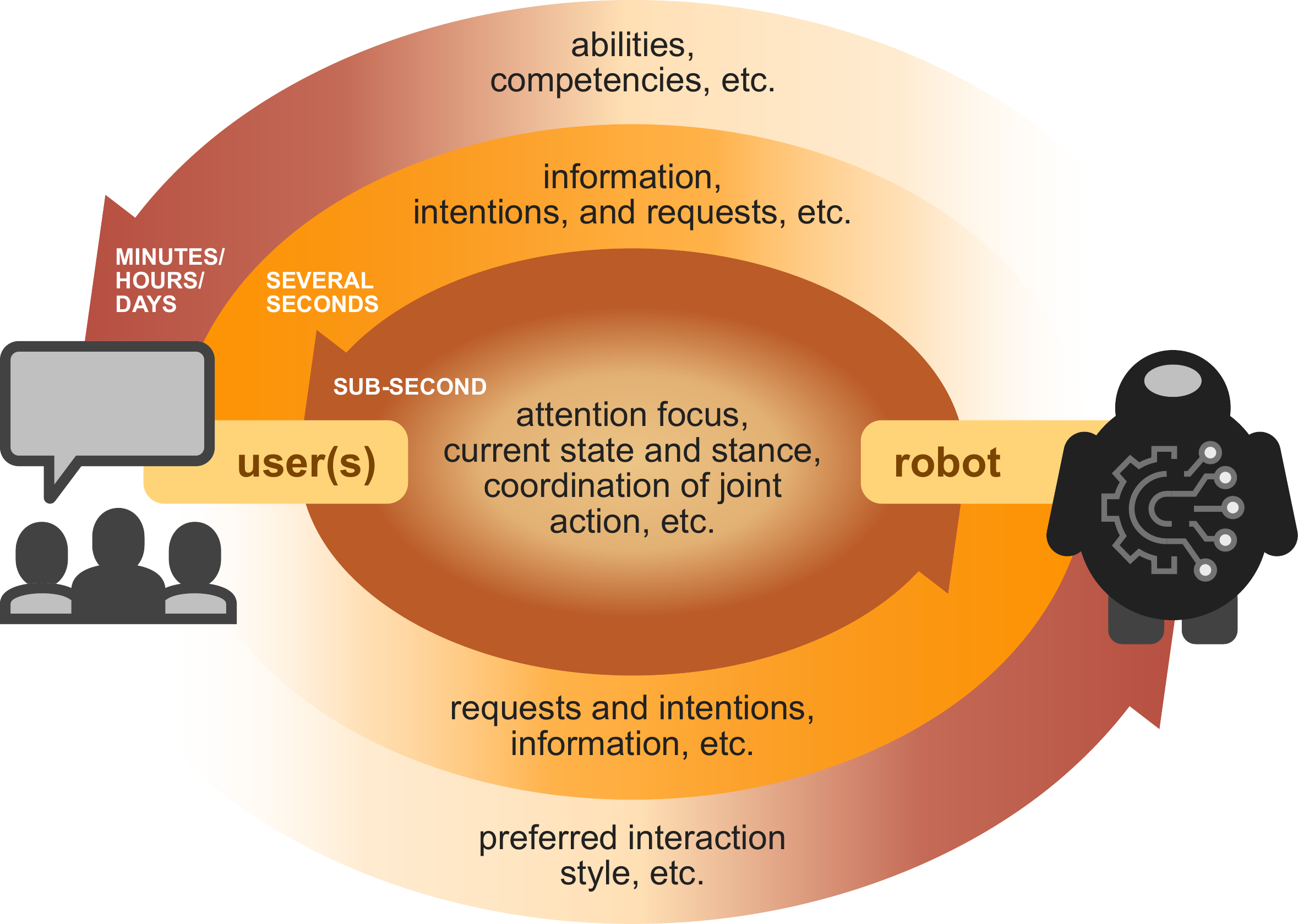}      
\caption{Speech-based communication between users and robots at three time scales. 
The audio signal at any moment may include information relating to all three time scales.}
    \label{fig:timescales}
\end{figure}

\medskip
\textbf{What is spoken communication, in essence?} It is not just audible text;
nor is it just transmitting packets of information back and forth
\cite{reddy1979conduit}.  Rather, in general:
\begin{enumerate} 
\item Spoken communication is a way for people to indicate their
  needs, desires, goals, and current state.  State includes
  internal state, such as stress level, level of interest, and overall
  emotional state, and also stance, such as attitudes and intentions
  regarding the current activity, teammate actions, and specific
  objects in the environment.
\item Spoken communication can relate to the open world, as it
  calls out objects of interest, signals upcoming actions, enables
  coordination with teammates, supports timely action, and so on.
\item Spoken communication can accompany actions and gestures to emphasize or disambiguate intentions.
\item Spoken communication serves interpersonal functions --- in
  motivating or guiding teammates, as well as in showing awareness of their
  contributions, their current state, their autonomy,  their value, and so on.
\item Spoken communication styles can portray diverse information
  about the individual, or robot, including its abilities, level of competence, desired interaction style, and so on.
\item Finally, spoken communication operates at various timescales, as suggested by Figure \ref{fig:timescales}.  
Of course, the robot's audio output should accurately give the
user information on the robot's current knowledge state, needs, and
intentions, and conversely the robot should understand instructions from the
user.  Such calmly paced utterances and responses have been the
primary focus of past research.  Yet robots often also need to be able to interact
swiftly with the user, enabling the direction of attention, exploitation of rapid dialogue
routines, and tight coordination of joint action.  Moreover, 
robots should be able to use spoken interaction when establishing long-term expectations.
In one direction, the robot's voice and turn-taking style should enable
the user to infer the robot's ``personality,'' including what the robot
is capable of and how it can best be interacted with.  In the other direction, the robot should be able to infer, from the user's
speaking style and interaction style, how this specific user likes to
interact, and to adjust its behavior parameters accordingly.  
\end{enumerate}

Not every robot needs competence in all these functions of speech.  If
a robot's job is just to pull weeds, it may need speech only for
receiving simple commands and providing simple status reports. But to
fully exploit the power of speech, roboticists will need to endow
their creations with new representations and new functionality.

\medskip
We offer the following contributions:
\begin{itemize}
    \item A brief overview of the critical challenges associated with spoken language interaction with robots, spanning user experience design, robustness, adaptability, infrastructure, speech processing, behavior signal analysis, language understanding, dialogue, human communication dynamics, language generation, and speech synthesis.
    \item A set of concrete recommendations to researchers and program managers alike that we believe will accelerate progress in this area. 
\end{itemize}

\medskip
This report is organized as follows. The following sections describe research issues and recommendations associated with
user experience design (Section~\ref{HF}), robustness and adaptability (Section~\ref{robustness}), technical infrastructure (Section~\ref{infra}), audio and speech processing, speech recognition, and behavior signal analysis (Section~\ref{asp}), language understanding (Section~\ref{lu}), dialogue and human communication dynamics (Section~\ref{dialogue}), and language generation and speech synthesis (Section~\ref{nlgtts}). We conclude the report with a list of recommendations tailored to funding agencies; these aim for improving the intellectual merit and broader impact of research in spoken language interaction with robots (Section 9). 

%% file: human_factors.tex
\section{User Experience Design}\label{HF}

At the top level, we can say that there are the three driving forces
that underlie most robotics research, and in
particular, most projects relating to speech for robots: the visions,
the technologies, and the needs.  While many technical challenges
remain, and we still need the inspiring visions, 
the field is now reaching the point where focus on the
needs --- the human needs --- should become the main driver.  
This section discusses what this implies, organized around four broad
 recommendations.

\reccox{U1}{Focus on language not only as a way to achieve human-like
behaviors, but also as a way to support limited but highly usable
communications abilities.} 

From the earliest days, an inspiration for robotics research 
has been the creation of 
human-like artifacts.  However, experience with many user interfaces 
has shown that aiming to emulate a human too closely is often a
recipe for failure \cite{Balentine2007}.  
Grand ambitions are good, but we also need to 
focus on engineering spoken interaction capabilities to maximize
usability and utility.  Simple, minimal interaction styles can even
be natural, in their own way, for people.  
Empirically, even when designers aim to support natural
``conversational" interaction, users often resort to formulaic language
and focus on a handful of interaction routines that reliably work for
them \cite{moore2017spoken}.

\reccox{U2}{Deliberately engineer user perceptions and expectations.}

People invariably form mental models of the artifacts they interact
with.  These mental models help them to predict what these artifacts
are capable of and how best to interact with them. Without guidance,
users can easily form misguided mental models, based for example on
the interactions seen with robots in science fiction movies. But
designers can help users form more accurate mental models of robots by
choosing appropriate visual appearances, selecting appropriate voices,
and implementing appropriate behavioral competencies.  Using both
first impressions and accumulated experience, users can thus come to
feel comfortable in dealing with a robot.

A complicating factor here is that the state of the art in speech and
robotics today is uneven: some components perform impressively, while
others lag.  In implemented systems, lack of coherence can be
confusing.  Obviously this implies the need for more work on the
deficient modules, but varying levels of ability will always be a
problem in real robots.  Accordingly, the abilities exposed to users
may need to be deliberately limited \cite{Moore2017}, to avoid giving
an exaggerated perception of competence that can mislead users
regarding how to behave and what to expect.  
More generally, designers need to avoid possible ``habitability
gaps'' \cite{Philips2006}, where usability drops as flexibility
increases.  This can be an instance of the ``uncanny valley'' effect, in
which a near human-looking artifact (such as a humanoid robot) can
trigger feelings of eeriness and repulsion \cite{Mori1970}.  
Of course there can be
trade-offs between attracting users to engage in the first place and
enabling truly effective interaction \cite{luger2016like}.

\reccox{U3}{Work to better characterize the list of communicative competencies
most needed for robots in various scenarios.}  

Today some research in speech for robotics follows well-worn paths,
extending trajectories inspired by classic taxonomies of language and
behavior.  These topics and issues are not, however, always the most
practically important for human-robot interaction. As suggested above
in Figure \ref{fig:apps} and elaborated below, there are many diverse communicative
behaviors worth modeling. In particular, we see the need to model
spoken interaction at rapid time scales, and to model it as
centrally involving social intentions.  These abilities do not
represent merely nice-to-have features; rather they provide the very
foundation of spoken interaction.

More generally, we see value in occasionally stepping back from the bustle
at the speech technology research forefront, to observe how people
actually communicate and what is most important for communicative
success.  This will enable us to thoughtfully determine what aspects of
speech are truly the most important, across diverse scenarios, and
thus to prioritize what to work on for the sake of maximizing the
effectiveness of future speech-capable robots.

\reccox{U4}{Design for use in multi-party and team situations.}

Today human-robot interaction is generally designed to support single users. 
While we do not wish to prejudge what new priorities might be
uncovered by such diligent observation work, it is clear that one priority must be support for
multi-party interaction.  
Many robots will function in environments with more than one person.  
These may include members of a team tasked to jointly work with the robot, 
complete bystanders, or anything in between.  Moreover the roles
of the humans may change over time.  

Support for the ability to deal with multiple humans in the environment has 
implications for all components of robots, including audio processing, 
speech recognition, speaker diarization, language understanding, computer vision,
situation planning, action planning, and speech generation and synthesis.
In particular, a robot
must be able to detect whether it is the intended recipient of some
communication or not~\cite{chen2011tale}.
Multi-party interaction with robots has already been demonstrated
in some situations, for example with the 
Furhat~\cite{al2012furhat} and Directions~\cite{bohus2014directions} robots,
and with Waseda's Facilitation robot, used to help people who might
otherwise get left behind in complex multi-party 
conversations~\cite{matsuyama2015four}, but enabling multi-part interaction
more generally remains a challenge.

%% file: robustness-adaptability.tex
\section{Robustness and Adaptability}\label{robustness}

In robotics research, like many other fields, successful demos are
celebrated: the demonstration of a new technology that brings us
closer to a noble vision of the future can be a source of great inspiration. 
However, success in demos is not very predictive of success in
deployment, especially since most demos only illustrate an  ideal case.  
However,  the field is maturing, and, as we increasingly
target solid experimental validation of capabilities and real
deployments in the open world,  considerations of robustness and adaptability are
becoming ever more essential.  This section recommends some strategic
directions towards improving robustness and adaptability.

\reccox{R1}{Include partially redundant functionality.} 

Human interaction is highly redundant, with the same message often
being conveyed by words, prosody, gaze, posture, pose, facial
expressions, hand gestures, actions, and so on
\cite{admoni2017social,gaschler2012social,ward-book}.
While robots can perform well in demos with only one of these
functions, this is only true when both the environment and the user
are tightly constrained.  Adding competence with these other modalities,
beyond just the words alone, can
contribute to robustness.  Achieving this requires better scientific
understanding of these aspects of behavior, more shareable software
modules, more explorations of utility for various use cases, and more
work on cross-modality integration, both for speech understanding and
for speech and behavior synthesis.

\reccox{R2}{Make components robust to uncertainty.} 

Demos can be staged so that the robot has complete knowledge of all
relevant aspects of the situation, but in open worlds such knowledge is not
possible.
To give three examples: First, for intent recognition, a developer
cannot assume that a robot can ever have a 100\% correct understanding
of the user's goals and intents; rather it will invariably need to
maintain a distribution of belief over multiple hypotheses.  Second, a
developer cannot treat the interface between the language
understanding module and the response planning module as a
single-predicate symbolic representation, given the inevitability of
alternative possible real-world referents and meanings that might be
ambiguously or deliberately bundled up in any user utterance.  Third,
a dialogue manager cannot be a simple finite state machine, as robots
need to track multiple dimensions and facets of the current situation,
typically none of which can be identified with full confidence.

It is easy to say that achieving robustness mandates that each
component constantly tracks multiple hypotheses and maintains a
probability estimate for each.  However, doing so involves many
challenges.  One of these is the need for something of a change of
mindset: we may need to accept the fact that simple, understandable,
inspectable representations may not be generally adequate.  Another is
that, even when we know how to make one or two components
probabilistic, it often remains difficult to integrate them, let alone
to design a software architecture for a robot that operates entirely
in this way.

One possible way out is end-to-end modeling, in which module
boundaries are erased, everything is jointly optimized, and all
mappings are learned directly from data. However, full end-to-end
training will likely never be possible for robots, as we will never
have enough training data for all the complex and diverse tasks that
robots need to do.

Thus we need research on ways to move towards systems that can
represent and model uncertainty throughout all components. As
suggested by the recent explosive advances in spoken dialogue systems
--- in large part attributable to the adoption of this perspective
\cite{sarikaya2017overview}, along with innovations in the design of
suitably simple representations and the judicious application of
machine learning from large data --- the potential benefits are large. 

\reccox{R3}{Focus not only on improving better core components, but also on
  cross-cutting issues.}

From a robot designer's perspective, it would be convenient if natural
language could be a simple add-on to an existing robot control
architecture.  Thus a naive approach to getting natural language onto
robots is to add an automatic speech recognizer (ASR) whose output is
fed into a finite state machine (FSM) that transitions to the
appropriate next state and simultaneously selects the next output to
send to the text-to-speech (TTS) synthesizer.  While such systems can
often meet the basic interaction needs in simple robotic applications
--- such as greeting a customer, prompting for a simple request, and
following simple instructions --- this simple ASR+FSM+TTS architecture is
not sufficient for more natural interactions, especially not in
``open worlds.''  Unfortunately, research effort has tended to gravitate
to these familiar components.  While we can, and should, add
components --- such as a natural language understanding (NLU) module,
a dialogue manager (DM), and a natural language generator (NLG) ---
there are many issues that fall through the cracks of such an architecture.
These include ambiguity, grounding, social adeptness, prosody, and
adaptation, some of which we discuss further below, and these, rather
than further refinements to pure core technologies, are often of the
first importance.

\reccox{R4}{Make every component able to support realtime responsiveness.}

Current dialogue-capable robots offer only slow-paced, turn-based
interactions, with few exceptions \cite{skantze2015exploring}.  This
is now due less to actual processing time requirements, than to the
architectures of our systems.  In particular, it is far easier to
build a system component if that component can delay the start of
processing until the upstream model delivers a fully fleshed-out chunk
of information.  For example, it is easier to build a recognizer that
waits until the user has produced a full turn and definitively
ended it.  Yet robots that operate on their own timescale 
can get out of synch with what the user is thinking, saying, and doing.  
Robots in general need to be responsive: to operate in real time.

Thus they will, in general, require incrementality in every component.
That is, each component will need to process data as a continuous
flow, incrementally and asynchronously updating its output
representations or probability estimates as new information comes in.
Incrementality in spoken dialogue has been an active area of research,
with work on incremental turn management, speech recognition,
semantics, dialogue management, language generation, and speech
synthesis
\cite{Baumann2017,2300868,Hough2015,Kennington2017a,kousidis2014multimodal}.
There are also general abstract models and toolkits for incrementality
\cite{Baumann2012,thilomichael2019a,Schlangen2011}, but much remains
to be done to make incrementality generally available to developers
needing to add dialogue capabilities to robots \cite{Kennington2020}.

Moreover, software for robots will generally need to model time
explicitly, in every component.  On the input side, robots have many
sources of sensory input beyond the speech signal. These include
cameras to take in scene information (e.g., for navigation or for
identifying objects, people, or gestures), laser scanners to generate
point clouds (e.g., for avoiding collisions), ``introspection'' on the
internal states of the robot (e.g., state information about movement,
location, or knowledge), as well as infrared sensors and GPS
information.  Different sensors operate at different sampling rates,
and the downstream processes --- for example speech recognition and
object detection --- have different processing speeds.  These will
produce different delays between events in the world and the time they
are recognized.  For example, if a user points to an object and then
to a location while saying \emph{put that there}, then the robot must
appropriately fuse the information from the speech and visual inputs.
The need for proper handling of time applies to all modules and
aspects of processing, including perception, planning, action
execution, speech recognition, gesture recognition, and speech and
gesture production.   The Microsoft Research Platform for Situated Intelligence
\cite{Bohus2017} provides mechanisms for this, but issues of
synchronization and temporal alignment still bring many challenges.  

Similarly on the output side: spoken output must be timed and
synchronized in concert with actions in other modalities, as discussed
later.  With advances in the synthesis of non-verbal actions, the need
here is becoming more pressing.  For example, a gesture at the wrong
time can be far worse than no gesture at all, and small variations in
the timing of responses to questions have large effects in the
interpretation of what people really mean by those responses
\citep{boltz05,kitaoka09,roberts13}.  While we understand some aspects of
these issues, so far our knowledge is mostly isolated to
specific aspects of specific dialogue acts in specific contexts.  In
general, there is a need for more general models of how to time and
align multimodal actions. 

\reccox{R5}{Make systems and components adaptable to users.} 

Every successful robot application today involves careful engineering
to make it work for a specific user population. This is especially
true for speech interfaces.  This process is expensive and slow, so we
need to face up to the challenges of making robots able to readily adapt,
either to groups or to
specific users and teams.  This adaptation might be partly automatic,
partly based on small sets of training data, and partly handled by
exposing parameters that developers can easily adjust.  Adaptation is
also necessary as a way to overcome whatever biases might exist in
training data, since no training set will ever precisely represent the desired
robot behaviors.

Further, even within a target population, each user is an individual,
and individuals will differ in age, gender, dialect, domain
expertise, task knowledge, familiarity with the robot, and so on.  One
particular open challenge is that of adapting to the user's
interaction style preferences. Today our understanding of interaction
style differences is limited.  We do know, for example, that in
multimodal interaction some people tend to make the pointing gesture
in synchrony with the deictic, as in {\it put it {\bf \textit{there}},} while
others tend to point after the word {\it there}
\cite{oviatt1997integration}.  We know that some prefer swift turn
taking with frequent overlaps, while others prefer to wait until the
other is silent before speaking \cite{tannen-meant}. We know that some
people like to explain things by a brief low-pitch monologue, while
others tend to explain by interleaving short pieces of an explanation
with frequent checks that the listener is following \cite{ward-book}.
We know that dialogue partners often accommodate to resemble the
interlocutor's behavior in terms of surface-level features such as pitch
height or speaking rate \cite{giles-mulac,levitan12}.  In addition
there is a rich folk vocabulary for describing interaction styles ---
including terms like stiff, withdrawn, shy, domineering, nerdy,
oblivious, goofy, chatterbox, quick-witted, lively, and supportive ---
reflecting the importance of these styles for success in interactions.
In the past, interaction style differences have not been a burning
issue, since most people are able, entirely subconsciously, to model
and adapt to the interaction styles of their teammates.  Existing
explorations in interaction styles and tendencies 
\cite{geertzen15,grothendieck11,hoegen19,hudry13,metcalf2019mirroring,ranganath-jurafsky}
need to be extended to model more aspects of behavior.
Beyond basic research, we need to develop ways for robots to
effectively embody plausible and consistent interaction styles, and
that can adapt to work with  different interaction styles.

In general, if robots are to become effective partners, we need
better models of the relevant dimensions of human variation, and of
how to adjust behavior to work well with diverse human partners.

\reccox{R6}{Develop new ways to make components more reusable across tasks and domains.}   

Robots need to be able to adapt to new tasks and domains. Linked to
system-level adaptability and reusability, there is also the question
of component-level adaptability. Developers of software components have
a general strategic choice of aiming to optimize performance for a
specific task by a specific robot, or of aiming to create reusable
components that can be plugged into any architecture and used for any
task.  This is an essential tension, but one that can be partially
alleviated.  One direction is to investigate how to best define
inter-component interfaces, either APIs or intermediate
representations, to enable better information fusion and thus better
decisions.  A second direction is to develop improved ways for rapid
adaptation to new contexts of use~\cite{tan-etal-2019-learning}, to enable the creation of
components that are simultaneously high-performing and highly
reusable. This may involve pre-training on massive data sets, with
mechanisms for easily and robustly ablating or adapting the models to
perform well on specific small domains, including, for some
experimental purposes, exceedingly narrow domains.

%% file: infrastructure.tex
\section{Infrastructure}\label{infra}

Research in  spoken dialogue for robots has high barriers to entry.
An ever-present problem is the intellectually demanding nature of conducting such interdisciplinary research: speech researchers must have access to a robot (or virtual robot), and robotics researchers must have access to some form of speech processing. 
To conduct research in this area requires mastering knowledge about  robot platforms and spoken dialogue frameworks, including individual components of both. Significant effort is required to create systems that  work, even minimally, for example  because individual components, such as automatic speech recognition, even when well-tested in other domains, often don’t transfer well to robots. This section describes infrastructure needs to conduct research on spoken language interaction with robots.

While the ultimate robot dialogue architecture will be 
complex and satisfy all
the requirements discussed above, in the meantime, the community needs
systems that make it easier for newcomers to get started, in the form
of accessible robotic platforms that come coupled with accessible
spoken dialogue systems.
There have been many notable efforts that partially address this need, of which we can mention only a few. 

\paragraph{Robotics Technologies}
The Robot Operating System (ROS) is a well-established platform for
robotic systems. 
It enables low-cost robots built on ROS such as
Duckietown (Ground and Air), MIT's Race Car platform, UW's MuSHR platform, and Artzi’s drone platform. 
Other efforts are not necessarily tied to ROS but endeavor to make
conducting robot research accessible: Microsoft
Research’s AirSim, Intel Research’s CARLA, Facebook Research’s AI
Habitat, AI2’s Thor, Nvidia’s Isaac Platform, 
Semio Arora, the Furhat Robot, Anki Cozmo and Vector, and Misty Robotics's Misty II.

Modern architectures for human-robot interaction lack the ability
to capture context with humans. If they do exist, they are usually
in standalone models separate of the ROS software stack.
As middleware, ROS does permit the capability to operate at latencies
supporting realtime human interaction. In practice, however, 
researchers have generally made limiting and sometimes inappropriate
assumptions about human interaction, in particular regarding highly time-sensitive tasks of
the kind common in social robotics. The tools
presently available using ROS for human-robot interaction 
can serve as a possible starting
point for speech researchers interested in getting started in
robotics.  In particular, users can get started with ROS before buying
any hardware since there are many virtual robot simulation packages.

\paragraph{Speech Technologies}
On the speech technology side, there are many available components,
both commercial and open-source.  For speech recognition, these
include Kaldi, Sphinx, and Deep Speech 2, and 
proprietary cloud-based systems
including Google ASR, Azure, and Alexa, although of course the lags of
cloud-based speech recognizers are prohibitive for many robotics
applications. 
The enduring problem here is that the openly available speech 
technologies generally have steep learning curves to 
getting started, while the proprietary commercial 
technologies do not support control of various components 
(e.g., controlling for vocabulary or retraining the 
language model to a robot task domain). 
For natural language understanding, there is Microsoft's
LUIS, Rasa, and PolyAI.  However, for reasons already noted, none of
these components are truly ``robot-ready.''
There are also a number of notable integrated frameworks and toolkits,
several of which have been successfully used in robots, including
InproTK, OpenDial, IrisTK, DIARC, RETICO, Plato, and the ICT Virtual Human Toolkit. These technologies provide evidence
that a fully integrated platform is possible.

\reccox{I1}{Create and distribute one or more minimal speech and dialogue-capable robot systems.} 

Although there are many tools and components, the easy-to-use ones are few and far between.  Ideally there should be a basic dialogue-capable robot that people
could simply buy and use out of the box.  Of course, what such a robot
should include is not obvious, given the many ranges of desired uses.
They range from fun, for hobbyists, to serious, for conducting interesting extensions and experiments.  They could support individual use 
or serve as a shared platform to help bring together researchers from
robotics, speech, social interaction, computer vision, and so on. Possibilities span simply concatenated modules to systems engineered
to support data capture, replay,
visualization, informative experimentation, and performance analysis.
They could vary from highly inclusive to exceedingly minimal, in the extreme case consisting of just a recognizer, synthesizer, and robot, with everything
else to be designed (or kludged) for the intended use.  

While no single solution will serve all needs, the availability of
a basic dialogue-capable robot would greatly increase the number of researchers
able to contribute to this area. To improve the likelihood
of wide adoption, the platform should have an extensible
simulated version compatible with ROS. In fact, a ROS-friendly
simulated platform would be free for users to try out
without the immediate costs of hardware. A critical step towards
identifying worthy platforms requires considering
the capabilities that robots will need to support
spoken language interaction with robots, described below.

\paragraph{Core Capabilities}

\begin{figure}[t]
 \centering
 \includegraphics[width=\columnwidth]{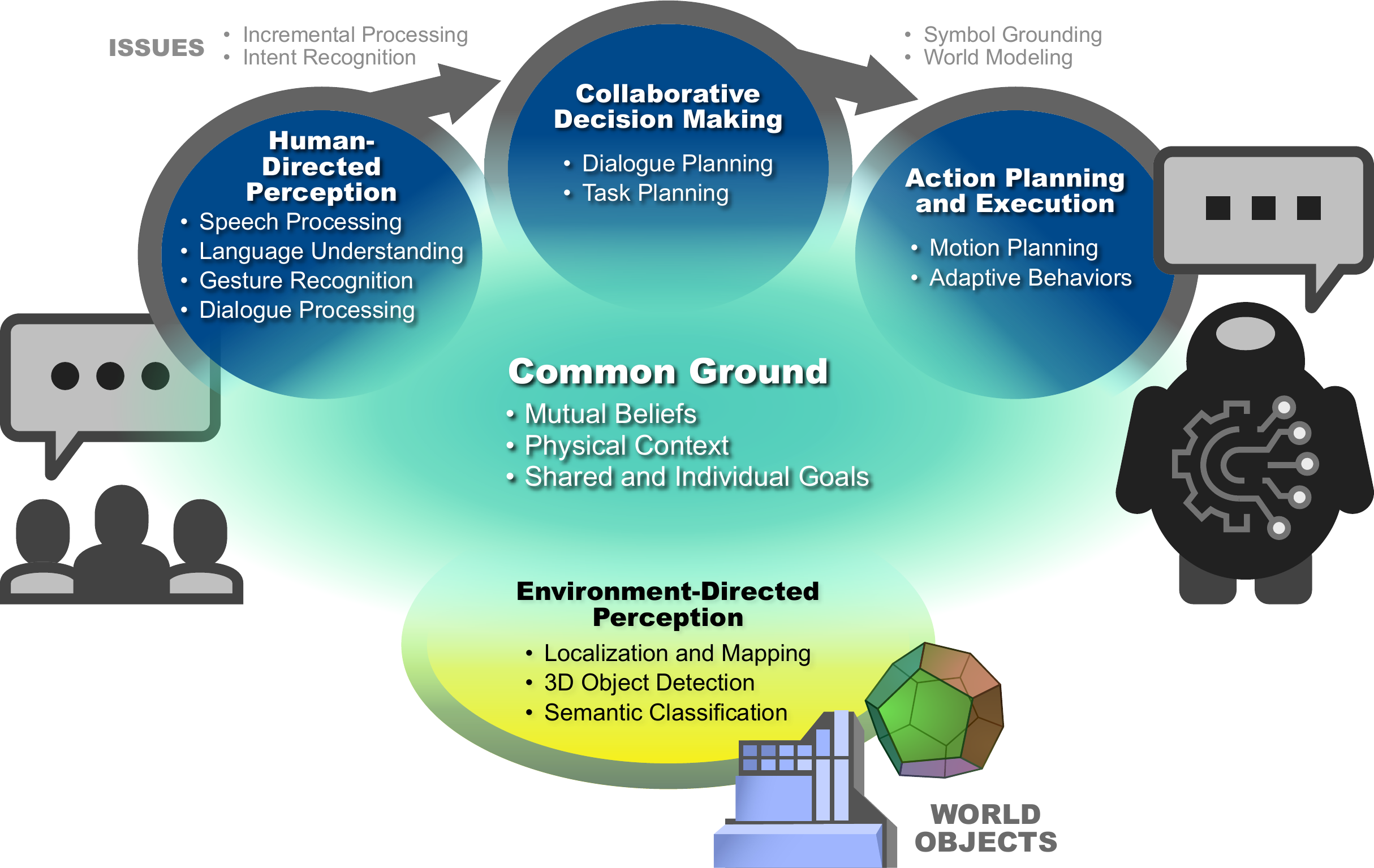}
 \caption{Core capabilities for robust spoken language interaction with robots include monitoring the human or humans a robot is interacting with and its physical surroundings, all of which contribute to common ground.}
 \label{fig:arch}
\end{figure}

Robots will need several features to converse
with humans in a natural way. 
Figure~\ref{fig:arch} presents
four primary capabilities that contribute towards common ground between humans and robots: (1) Human-Directed Perception, (2) Collaborative Decision Making, (3) Action Planning and Execution, and (4) Environment-Directed Perception. 
This diagram serves as a high-level 
overview of how robot intelligence architectures 
break down the problem of converting all of a robot's inputs 
into meaning representations of the physical world and 
actions that we want a robot to take, 
goals that we want to specify, 
constraints that we want them to satisfy, 
or facts about the world that we want to convey.
The next four paragraphs discuss issues related to these four capabilities.

For robust spoken language communication with robots, the robot must
first perceive the human or humans that are seeking to interact
with it. Software supporting speech processing extracts
sequences of words, along with audio-spectral data such as prosody,
from raw audio signals produced by human speech. Components
for language understanding convert word sequences into
semantic representations that convey the relationship between words
in the speech, determine a task intention, and attempt to ground
the intention in the robot's immediate physical context. 
Dialogue processing determines if the robot has produced
an initial understanding of what was said. For example, if the speech is
too noisy, it could initiate a clarification (e.g., request the user to repeat).
For these processes to work in real time and support fluid human
communication dynamics, they should be running incrementally
as data is fed into the robot's input streams. Ultimately,
the robot's sensing of humans will need to be distilled into one or more intents that it can recognize. 

Intents observed by the robot contribute to how the human(s) and robot
are to make decisions. These will often be joint, in the sense that
achieving common ground will require coordination from both
entities. Dialogue planning policies dictate how a back-and-forth
dialogue would play out given the robot's observation of an 
intent from a human,
a notion of the current context (physical, task-oriented, audio, and otherwise), and any shared
history. While some actions will be verbal (such as replies
or requests for clarification), some require broader task planning
so the robot can perform physical actions in its surroundings.
Dialogue planning also accepts information from the downstream
robot actions to report on successes, failures, and errors
to humans when appropriate.

Human intents must be encoded as goals (with corresponding
constraints) so that they 
can be mapped to robot actions. One such process, symbol grounding,
attempts to associate intents to the robot's environment. 
The robot's actions are determined by the adaptive behaviors it can do (i.e., behaviors that change based on the context, such as
performing an exploration task or following another moving agent).
Motion planning dictates the precise movements the robot makes
based on what it senses from world modeling.
Based on the current context or in reply to a human,
response generation allows the robot to verbally or non-verbally
provide output to complete the bi-directional communication. 

Without an environment model, the robot would not be able 
to meaningfully refer to task-relevant objects in the physical world. 
For a robot to contextualize speech from a human, it must first
understand where it is in the world, which is achieved by 
localization and mapping (e.g., SLAM). Computer vision
technologies from object detection and recognition contribute
to a robot's ability to perform 3D object detection.
Finally, a broader environment model is derived from 
semantic mapping that accepts observations from a set of 
processes representing semantic classification.  

It is not the intent of Figure \ref{fig:arch} to serve as a definitive and conclusive answer for how to design a robot intelligence architecture capable of performing bi-directional communication with humans through speech, but rather it serves as a representative framework that situates many important problems in spoken language interaction with robots.  Many subtleties, across
the range of lower-level processes in audio, speech, 
and robotics processing, are omitted for clarity.
From this diagram we can also see how high error rates in 
speech and audio processing or deficiencies in how dialogue 
is tracked over time could significantly impair how a 
robot reasons about the world and selects actions to perform various tasks.

\paragraph{Common Ground}

All four capabilities described above contribute to building a 
common ground representation. All signals interpreted by the robot
contribute to its building of a set of inferred mutual beliefs with one
or more humans. Meanwhile, the robot must also build a notion
of its physical context, which consumes non-linguistic auditory, visual, 
and tactile inputs into observations to infer 
distributions about metric and semantic models 
of the environment.    

\reccox{I2}{Update the representation of a robot's physical surroundings continuously.}

Successful interactions between humans and robots will require not only robust speech processing, 
but also the ability to quickly and
accurately contextualize spoken language dynamically with
interpretations from the robot's immediate physical context. 
These interactions will be open world in that they 
will take place in unconstrained environments and 
rely on continuous properties like space and time~\cite{bohus2009models}.
This demands an ability to quickly assess multiple sources
of uncertainty that will inform a robot's ability to perform
actions. Tasks that have proven successful in open world
contexts include learning from
demonstration~\cite{argall2009survey} and interactive 
task learning for robots~\cite{chai2018}, though spoken language has not 
traditionally been the primary source 
of communication.

Open world interactions form the
robot's physically situated context~\cite{kruijff2007situated}: (1) perception, which includes 
sensor readings from visual and occupancy-based sensors,
(2) knowledge about pre-defined or 
recently acquired knowledge about 
the physical world or relations within it, 
(3) spatial reasoning
about relationships between the robot, nearby objects, and other agents, and (4)
temporal reasoning about how things in the robot's world change
over time. Although speech recognition will often be an inadequate source of information~\cite{marge19-tiis},
these other sources will provide crucial insight on possible
misunderstandings and therefore require the development of
novel representations that support continuous reasoning in the 
physical world. Compounding this problem is the diversity of
sensors: they range from microphones (both headset-based
and array-based) to a robot's onboard cameras, depth sensors, laser scanners, 
haptic and proprioceptive sensors, and other peripherals. 
Ensemble machine learning
methods and representations will likely be required for
robots to process speech and other sensors while
reasoning in the physical world.

%% file: speech.tex
\section{Audio and Speech Processing,  Speech Recognition, and Behavior Signal Analysis}\label{asp}

Roboticists looking to exploit speech recognition today face numerous challenges.  
While everyone is  familiar with high-performing speech recognition, 
approaching or exceeding human performance in some cases, making use
of the technology for robots is very hard. As noted above in Section \ref{infra}, speech researchers must have access to a robot (or virtual robot), and be willing to develop the knowledge required to work with robot platforms.  This section recommends the development of resources, speech-related technologies and the use of paralinguistic information to aid in the development of a speech recognition system specifically for robots. 

\begin{figure}
    \centering
    \includegraphics[width=4in]{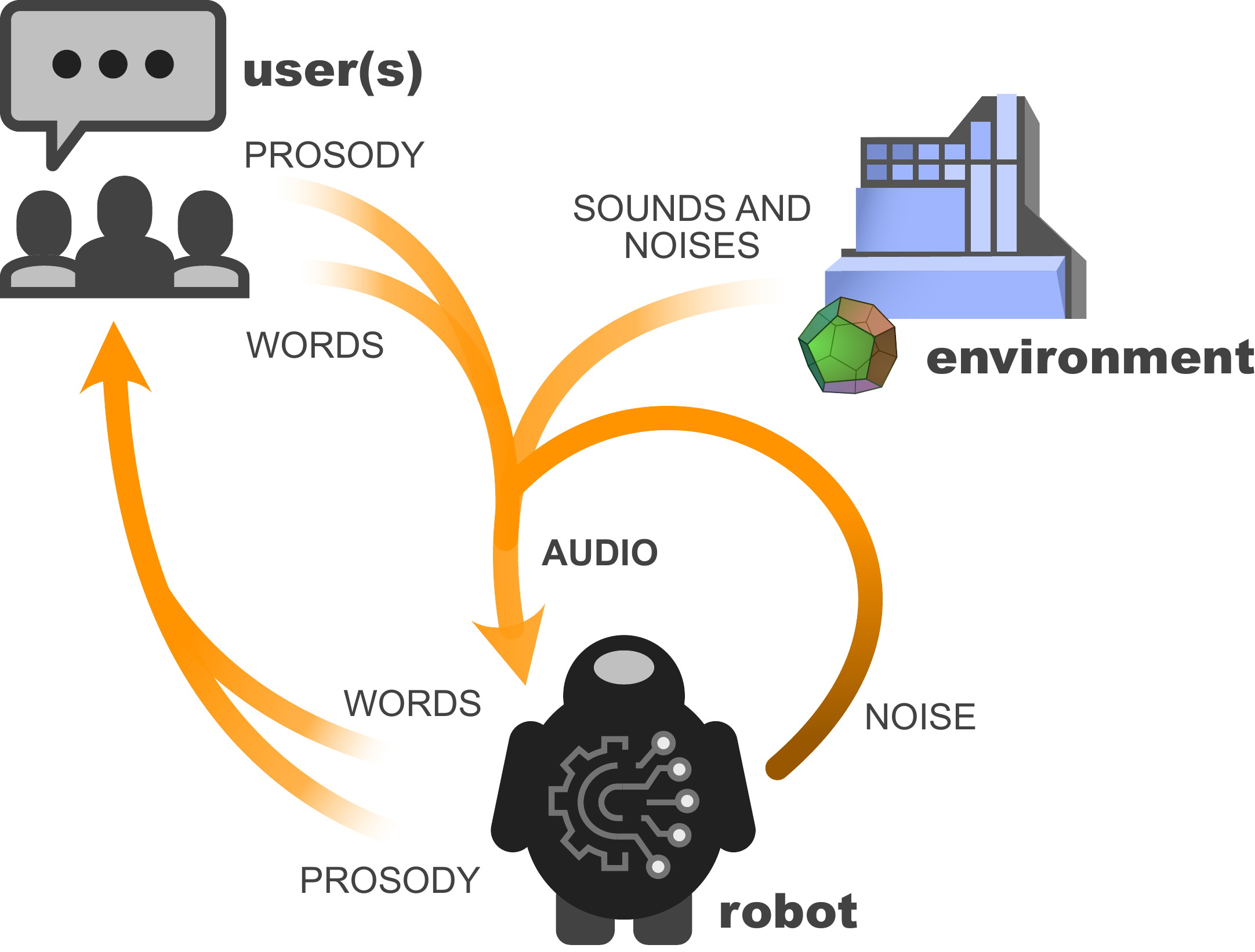}
    \caption{A robot's audio input will be a mixture of lexical and prosodic information from the user, plus meaningful sounds from the environment and the robot's own speech output, plus noise from the environment and its own motors. Disentangling these is a huge challenge.}
    \label{fig:audio-env}
\end{figure}

\reccox{A1} {Develop the following shared resources: general toolkits for front-end audio processing,  a database of robot-directed speech, and a challenge task on speech recognition for robot-directed speech.}

The first set of challenges speech for robotics faces relates to the acoustic environments where robots operate,
which are often very complex, as suggested by Figure \ref{fig:audio-env}. Even for a robot 
in a quiet, non-reverberant room with only a single user present, the sounds of the robot's own motors and speech
will be present.  Multiple speakers bring a further challenge.  Thus, prerequisite to speech 
recognition, there is a need to separate out the speech signal(s) from the confounding contributions
to the audio signal.  Aspects of this problem have been well-researched, with various techniques for source separation, both 
in software such as single-channel speech segregation~\cite{isik2016,mesgarani2019} and using hardware, such as microphone arrays (e.g.,~\cite{Chhetri2018}).  
Smart speakers, for example, do quite
well even in complex environments, thanks to intensively-tuned algorithms.
However, reusable general toolkits appear to not yet exist.  These should, ideally, be pre-trained
on massive datasets, but easily and robustly adaptable to specific contexts of use. In addition, current techniques will need to adjust to wheeled or bipedal robots that are moving around in their environment.

In the case of multi-human, multi-robot communication, a robust speaker diarization system will be needed to partition the speech signal into homogenous segments corresponding to each speaker.   While such partitioning is not necessarily needed before ASR is performed, it will be needed in order for the robot to understand what was said, and who said it.  The degree of difficulty of this task will depend on whether a supervised approach that allows for the system to be trained on speech from each speaker is possible or not.  There are several speaker diarization tools available including the open source tools available with pyAudioAnalysis \cite{giannakopoulos2015pyaudioanalysis} and Kaldi, and also others that are meant to work in conjunction with specific ASR systems such as LIUM, which integrates well with Sphinx, Microsoft’s VideoIndexer  which includes speaker diarization and ASR among other tools, and Google’s joint speech recognition and speaker diarization system \cite{shafey2019}.

One issue with cloud-based, pre-trained models like Google's speech-to-text system and Microsoft's Azure is latency and the need for an internet connection. Because the recognizers are cloud-based and not local to the robot, the latency in obtaining the recognition results is often too slow for practical and natural interaction with robots. Local speech recognizers can process faster, but the tradeoff is that locally trained models don't have as large of a vocabulary as the cloud-based models. Moreover, cloud-based recognizers are invariably trained using data sets and objective functions that are quite unlike those needed for robots.  Naturally, the performance obtained on speech to robots is much poorer.  Issues include most centrally the sorts of things that people tend to say to robots, and how they say them~\cite{marge-etal-2020-first}.  

To balance these tradeoffs, we suggest the development of, for example, a thousand-hour dataset of human-robot speech to enable training models with existing tools, and the evaluation of new techniques. Realistically, no single dataset could handle all the types of speech and situations needed for robotics, so any such corpus would need to be diverse, across multiple dimensions: speech directed to both humanoid and other mobile robots; in office, warehouse- or airport-sized spaces, and various outdoor environments; for a variety of tasks; for various user demographics; and for a variety of microphones including headset, on-board, and microphone-array.

\reccox{A2}{Better exploit context and expectations in speech recognition.}

Speech directed to robots will always be a challenge to recognize, but in partial compensation, the
context can be expected to be highly informative.  For example, if a robot has just started to move, the
probability of hearing words like {\it stop}, {\it wait}, or {\it no} will increase, and the 
probability of hearing words like {\it pick}, {\it lift}, and {\it explain} will decrease.
In other words, the robot can use its interpretation of the environment, 
task plan, and available actions to bias its language and speech understanding. 
Among the many ways to do this, perhaps the most practical is dynamic language modeling,
ideally involving a model able to map from the entire context to a probability distribution over 
all the words in the vocabulary, continuously updated.  Of course, speech recognizers need  APIs that support this. 

\reccox{A3}{Consider creating a speech recognition system focused on the issues encountered in robotics (e.g., speech segregation and speech in noise where the noise may come from the robot and/or the surrounding environment, possibly while the robot is moving).}

While many speech recognizers exist, none are  robot-ready.  In practice,
high error rates remain a key limitation for 
creating social robots.  While we expect that  
existing engines can be extended and adapted to work well
for robotics, it is also worth considering a branch of an existing open source
recognizer to create one specialized for robotics.  Reiterating some points made
earlier, this should be robust to high amounts of noise and multiple speakers, robust to spontaneous and fragmentary  
utterances, retrainable to perform well in narrow domains, incremental and fast, time-aware, 
and designed to work well with other components for environment tracking, prosody processing, multimodal input, and realtime output.

\reccox{A4}{Better exploit prosody, emotion, and mental state.} 

Speech includes both words and prosody.  Speech recognizers handle only the former, meaning
that much of the information in the speech signal is discarded.  Prosody comprises features of the speech input that are not governed by the phoneme sequences of the words said.  These include features of pitch, energy, rate, and voicing.  In many applications the lack of prosodic information is not
an issue: if the user wants to set an alarm or to get today's weather, it's enough to detect the words,
without worrying about whether the user is confused, preoccupied, distressed, unsure, or about how
the utterance relates to the user's goals or the temporal context.  Yet for robots, all these
aspects, and many more, can be critical.  In some cases, the prosody can matter more than the words: 
an {\it oops} can flag an embarrassing little mistake that can be ignored or a major surprise that 
requires everything to be replanned, and only the prosody may indicate the difference.  Configurations of prosodic features convey information of three main kinds: the paralinguistic, conveying user traits and states, the phonological, relating to the lexical and syntactic components of the message, and the pragmatic, relating to turn taking, topic structure, stance, and intention.  
Today it is easy to compute many prosodic features, and, given enough training data, to build
classifiers for any specific decision.  However, we would like tools that can extract prosodic 
information in real time and provide a continuous read-out of the results, in terms that
are directly useful as input for robot task planning.

In addition to recognizing what the user has said, an ultimate goal will be for robots to be able to use speech and language as behavioral signals to assess the user's current state: mental, physical, and otherwise.  There is considerable ongoing research in emotion recognition, sentiment analysis and mental health condition based on unimodal and multimodal systems using speech, language, facial gestures, and various biological signals (e.g., \cite{cummins2015}). Having robots take advantage of behavioral signal processing can be crucial in determining how to best be of assistance.  For example, when helping a child read, if the robot is able to ascertain that the child is getting frustrated, then it can change its approach, and continue to do so until it determines what will be most effective in helping the child learn.  As another example, a robot that is participating in elderly care might alert family members or a doctor if it is able to recognize negative symptoms in emotional state that could suggest the onset of depression.

\reccox{A5}{Use audio scene and event analysis to better understand the environment.}

Finally, not only is high-performing speech recognition needed and the determination of mental state desirable, robots also need to understand the sounds in their environment, which provide contextual information that can be crucial to their performance.  
As an example, consider a robot assisting first responders in a disaster relief setting. The robot should be able to sense sounds from vehicles and bystanders that could pose a danger to relief efforts and notify human teammates. 

%% file: lu.tex
\section{Language Understanding}\label{lu}

Language understanding focuses on interpreting communications
from human to robot into a machine-usable representation. 
While intelligent personal assistants primarily focus on
deriving intent from a user's speech to determine a system
request, robots must do more. 
Specifically, they must perform grounded (or situated)
language understanding, where success depends on correctly
interpreting language that refers to, or has meaning only
in the context of, some sensed physical environment. 
In this section, we describe current approaches to
grounded language understanding and offer recommendations
that leverage this context to improve spoken language interaction
with robots. 

Grounded language understanding is often necessary for robots 
to follow human-provided instructions. The 
task can be framed as the problem of interpreting a linguistic 
structure, possibly in the context of a perceived or 
\textit{a priori} known environment model, to produce a representation that is meaningful to a robot. In robots, 
interpretations drawn from language understanding 
can then be shared with a dialogue manager to determine actions, 
goals, or constraints to mission planners, motion planners, 
or planning and scheduling modules. Meanwhile, 
a language generation module can communicate a 
statement or question back to the human interacting with the robot.  
The task of grounded language acquisition goes hand-in-hand with
understanding, where the agent learns a correspondence
between some interpreted natural language text and the environment
with the goal of fulfilling one or more tasks. 
In practice, given the complex physical world in which 
robots operate, such models need to be learned \textit{in situ}, 
rather than predefined for all language understanding tasks.

Generally speaking, grounded language understanding seeks to 
infer a representation that associates natural language to 
a robot's environment model as perceived through 
sensing (e.g., vision, sound, haptics, etc.~\cite{thomason2016learning}). 
This environment model may contain spatial and semantic 
information that can be used to resolve the meaning of 
utterances where resulting interpretations relate in a meaningful
way to physical objects in the environment that humans 
and robots share. The ability to understand expressions 
that relate to or inform a robot's model of the 
environment is critical for collaborative robots. Such robots may
be required to follow sequences of 
navigation~\cite{anderson2018,chen2011learning,macmahon06a} 
or manipulation~\cite{chai2018,paul18a} instructions using spatial terms~\cite{paul2017grounding} or 
report back on observations in areas 
beyond a human teammate's line of sight. Several representations
have proven effective, such
as Generalized Grounding Graphs~\cite{tellex2011approaching,tellex2011understanding}, 
Distributed Correspondence Graphs~\cite{howard2014natural}, Dynamic Grounding Graphs~\cite{park2019efficient},
linear temporal logic propositions~\cite{boteanu2016verifiable,boteanu2017robotinitiatedinteraction}, combinatory categorial 
grammars~\cite{artzi13a,matuszek2013learning,thomason:jair20}, and neural networks~\cite{blukis19a, mei2016listen, misra17}.

There are a number of open problems in grounded language understanding.  
As the recent literature has shown, the field has made 
significant progress in the ability to efficiently infer 
accurate representations using machine learning~\cite{tellex2020robots}. 
Nevertheless, gaps still remain in our ability to 
(1) track user intention in spoken dialogue, 
(2) resolve complex spatial and temporal relationships 
between referents (i.e., objects 
and regions) in the physical environment, 
and (3) represent the world to most efficiently interpret 
the meaning of an instruction.  
Most evaluations 
have been conducted in simulated environments with clean 
and often well-formed written language inputs.
An inherent limitation to these methods is that 
they are not necessarily able to derive user intention 
from the shorter, less formally structured content 
in spoken language. However, the back-and-forth nature
of these communications can be exploited to make language
understanding more robust.

\reccox{L1}{Develop language understanding models for robots that resolve referential and other ambiguities in spoken dialogue.}

Not only will humans and robots be speaking
to each other, they will be co-present in a shared environment.
A touchstone task to assess progress 
can be to resolve referents in spoken dialogue
to accomplish tasks. This is because it satisfies the following three core 
requirements: an interaction that uses speech, 
grounding language to the environment, and bi-directionality. 
This challenge is non-trivial because humans 
and robots have mismatched levels of knowledge 
and abilities in recognizing speech, understanding language, 
perceiving the environment, and reasoning about goals, actions, 
and plans. 
To mediate perceptual differences, grounded language understanding
will need to go beyond single utterances and rely on 
collaborative discourse~\cite{liu2015aaai}.
If more efforts focus on resolving referential ambiguities 
in dialogue, we can identify coping strategies that mitigate this
fundamental difference. 
Another key challenge of grounded language understanding in spoken dialogue involves the timing demands: accurately inferring the meaning of an instruction 
in a partially known or highly noisy representation of the environment in real time.  
Efficiency is particularly important because without realtime language understanding algorithms (as described in Section~\ref{robustness}),
speech and language-based interaction becomes impractical.

\reccox{L2}{Develop methods to infer and represent more information about humans from robot sensors.}

For many robots today, a user is treated as just a disembodied source of 
speech input, or at best, 
an approximate image region with an estimated location 
and velocity. Sometimes such a simplistic representation is appropriate
(such as navigating in a space away from humans) 
and other times this representation is sorely inadequate 
(like social interaction with a hospital patient). 
In the latter case robots need to know more, both to 
properly understand what the user is saying and to appropriately
speak to them.  A robot should be able to infer what the 
user is doing, what
they are paying attention to, what their posture suggests they are likely to do next,
aspects of their energy level, current cognitive load level, 
and so on.  Doing this reliably requires advances in 
both speech and robotic perception. 

%% file: dialog_interactive.tex
\section{Dialogue and Human Communication Dynamics}\label{dialogue}

Dialogue can be broadly defined as interactive
communication (speech, gesture, gaze, etc.) between two or more
interlocutors~\cite{fong2003collaboration}. 
We wish to enable robots to converse with people, providing
a coherent, flowing, natural, and efficient user experience, but
this requires far more than just
incorporating the core language components. Realistic androids like ERICA~\cite{glas2016erica} have shown the promise of immersive realtime dialogue with robots, but there are still many active areas of research.
This section describes the need for realtime interactivity in 
robot-centric dialogue processing and identifies some key challenges. 

\reccox{D1}{Focus on highly interactive dialogue.}
 
Dialogue allows a robot to interpret a user's speech
and prompt for clarification if something is unclear.
At the same time, the robot can provide verbal status
updates of its progress on a task. 
For robots to build common ground  
with users, they will need to coordinate joint activities~\cite{bangerter2003navigating}, with speech playing a central role. 
Speech provides a direct signal to a robot
that a person is instructing it, and the acoustic interpretation
can serve as a source of information for managing dialogue and making decisions.

A critical 
capability for robots will be the ability to ask
for help or clarify when something is unclear. Such decision-making
will require determining the level of uncertainty about a robot's sources
of evidence when performing a task. Such sources are not limited to only
the human-robot interaction itself, but also the physical context,
including objects in the immediate physical surroundings 
and their properties, and actions and their immediate feasibility.

\paragraph{Improving Dialogue with Recovery Strategies}
With dialogue, a robot can mitigate failures to perfectly
understand language by asking for clarification using 
recovery strategies to get the conversation back on 
track. This can be achieved in multiple levels of human-robot
communication, whether they are related to the speech that 
was uttered~\cite{schlangen2004causes} or to 
the broader physical context~\cite{bohus2010challenges}. 
Clarifications can be broadly categorized as responding to non-understandings
and misunderstandings~\cite{hirst1994repairing}. Non-understandings
mean the robot would have no clear interpretation of what 
was said by a human, while with misunderstandings, the robot 
would arrive at an incorrect interpretation, though it could 
provide evidence of its reasoning if needed. While a great deal of recovery strategies exist in the dialogue literature, few are designed for human-robot communication.

\reccox{D2}{Explore the broad space of recovery strategies in
spoken language interaction with robots, including when and how.}

While there exists a well-established list of recovery strategies
for speech recognition (e.g., clarifying a word or 
phrase, or asking the user to rephrase 
their instruction~\cite{skantze2005exploring}), only a handful 
exist for clarifying physical context in a dialogue, 
such as resolving ambiguous or impossible-to-execute
instructions~\cite{marge19-tiis,thomason:jair20}. 
For robots, the space of recovery strategies expands to 
include not only speech but also actions involving gesture 
and movement. Early signaling of the need for recovery 
becomes possible, for example, with a raised eyebrow or a sudden
slowing of motion. 

\paragraph{Human Communication Dynamics in Realtime Interaction}
Realtime social interaction is a uniquely human ability.  
For many human abilities, AI is on track to approximate or exceed human
performance, but that is not the case with human communication dynamics. 
To reap the full benefit
of other advances, robots need to be able to work effectively with
humans, which requires research in this area. 
Realtime interaction is essential in joint-task situations
where time is of the essence, but also has more general value.  It is,
indeed, something that people often seek out.  Texting and emails have
their place, but if we want to get to know someone, negotiate plans,
make lasting decisions, get useful advice, resolve a workplace
issue, or have fun together, we usually seek a real-time spoken
interaction.  For robots to be widely useful and widely accepted, they
similarly need to master real-time interaction. 

However, this is currently beyond the state of the art.  To quote from \cite{ward-devault-aimag}, 
given the
broad acceptance of systems like Siri and Alexa, one might imagine the
problems of interaction are solved. But this is an illusion: in fact,
these systems rely on numerous clever ways of avoiding true
interaction. Their preferred style is to simply map one user input to
one system output, and they employ all sorts of stagecraft to guide
users into following a rigid interaction style.  Thus today most
interactive systems require tightly controlled user behavior. The
constraints are often implicit, relying on ways to set up expectation
and hints that lead the user to perform only a very limited set of
behaviors \cite{dog-book} to follow the intended track.  Such
constraints greatly simplify design and reduce the likelihood of
failures due to unplanned-for inputs.  However, designing around
narrow tracks of interaction has led system builders to adopt
impoverished models of interactive behavior, useful only for very
circumscribed scenarios.

In the research arena, researchers have shown how we can do better,
producing prototype systems with amazing responsiveness
\cite{devault-aamas14,gratch-engaging}.
For example, some researchers demonstrated a
system that could pick up on subtle indications of a student's
cognitive state and respond in ways that increased learning gains \cite{litman-uncertainty11b} and
others demonstrated how a
robot in dialogue could dynamically shape the user's attention
\cite{yu-bohus15}.  These illustrate that elements of true realtime
interaction are possible.  Moreover these are often highly valued:
experiments have shown that users interacting with systems (or
people) with better interaction skills trust them more, like them
more, and use them more \cite{fusaroli16}.

In general, we envision the creation of highly interactive systems. 
Imagine that you're moving heavy furniture, performing surgery, or
cooking with the aid of a robot.  You would want it to be alert,
aware, and good at coordinating actions, and this would require
competent realtime interaction. Borrowing again the words of 
\cite{ward-devault-aimag}, we think that in broad strokes, 
these systems will be characterized by low latency 
and natural timing, a deft sensitivity to
the multi-functional nature of communication, and flexibility about how
any given interaction unfolds. Their skill with interaction timing
will be manifest in the way they are attuned to and continuously
respond to their users with an array of realtime communicative signals. 

\reccox{D3}{Work to elucidate the fundamental questions in realtime
social interaction, both scientific and engineering.}

As our AI systems grow more
capable, the need for realtime social interaction competence will only
increase. Existing techniques are limited. Some involve
custom datasets, careful policy design, and intense engineering and
tuning, and these do not scale.  Others model only single dimensions
of interaction \cite{smith-chao}. At the same time, current deep
learning models, though they have worked so well in many areas of AI,
are not directly applicable to realtime, situated interaction.

%% file: synthesis.tex
\section{Language Generation and Speech Synthesis}\label{nlgtts}

So far our focus has largely been on communication from human to robot, but
robots also need to speak to humans.  Speech synthesizers were,
historically, designed to create an audio signal to intelligibly
encode any given sentence.  The target was read speech, in a neutral
tone. More recently, synthesizers have become able to produce speech
that is not only intelligible but also highly natural, and even
expressive in some ways.  While adequate  for some purposes, robots often
need more.

\reccox{S1}{Extend the pragmatic repertoire of speech synthesizers.}

Robots operate in real time and real space.  
Speech synthesis in this context needs access to the full
expressive power of spoken language.  For example,
consider the use of language to direct attention ({\it hey look!)},
convey uncertainty ({\it the red one?}), 
establish priorities ({\it help!}), or coordinate action ({\it ready
  \ldots go!}). With appropriate timing, voicing and prosody, such
phrases can be powerfully effective; without this, users may be confused or
slow to respond.  Or, for example, imagine a robot prefacing its next
movement with {\it okay, over behind that truck}.  Beyond the words, a
cooperative utterance may also convey  the robot's view of the likely
difficulty of moving behind the truck and its desire for follow-on
information about what it should do once it gets behind the truck.
Robot speech thus needs to be able not only to convey propositions and
speech acts, but to be able to simultaneously convey nuances of
information state, dialogue state, and stance.  Enabling robots to do
such things requires advances of several kinds.

To produce such richly informative outputs, speech synthesizers need
rich input: far more than just sequences of words. For the
above examples, effective speech synthesis would also need access to
information from the user model, environment and plan representations, 
and mission context representation.  Current software architectures for robots generally
do not expose such information: it may be buried down in some
component-specific internal data structures or parameter values.  To support
adequately expressive synthesis, we will need new ways to explicitly
represent and expose more of a robot's instantaneous internal state.

Beyond issues of speech synthesis, effective communication also
requires appropriate choice of words.  For some applications, a robot
may need to produce only one of a finite set of sentences, or use only
a finite set of templates.  However, effective language generation
remains a challenging problem \cite{gatt-krahmer}, especially for
robots, for reasons already noted.  The term ``speech generation'' is
sometimes used to indicate that speech synthesis and language
generation are essentially one, tightly integrated problem.  Today a
pipelined approach, with two separate modules --- concept-to-text and
text-to-speech (TTS) --- is the norm, but this is problematic
\cite{bulyko2001}, as the language and speech decisions are often
interdependent.  In some cases the interdependencies may be mild, for
example, for TTS systems used to produce good quality isolated
sentences based on read speech example.  Robots, however, require more
expressive power.
Today, end-to-end training may in principle solve this problem, but in
practice, the limited data associated with many human-robot
interaction scenarios will make this problematic. Instead, researchers
will need to pursue loosely coupled language and speech generation,
where the generated ``language'' comprises both text and control
signals for speech synthesizers. Of course, this means that speech
synthesis systems must be designed to allow such control. These
control signals will need not only to support wide expressive
abilities, but also the needs of a robot that performs actions in time
and space.  

The nature of these control signals is a question in
itself.  A particular challenge is that of appropriate prosodic
control signals. Clearly the use of punctuation marks is not enough.
For example, an exclamation point can indicate emphatic agreement
({\it exactly!}), enthusiasm ({\it let’s go!}) or urgency ({\it
  help!}).  Also, while an exclamation point can accurately indicate
emphasis in a short phrase ({\it over here!}), it would not be useful
in a sentence where meaning can differ depending on the emphasis
location, as in {\it this} versus {\it today} in: {\it We need to use
  this one today!}  It may be possible to learn from data an
appropriate set of prosodic control signals \cite{style-tokens}, but
it is not clear how ``style tokens'' or other methods for representing
tone in simple applications --- like audiobook synthesis and emulating
acted emotions --- can be extended to support speech
adequate for the here-and-now communicative functions \cite{ward-book}
that robots most need.  More work specifically targeting the speaking
needs of robots, perhaps involving adaptation methods and multi-domain
training, is needed.

Audience design is another major issue: robots need to produce speech
that is not only clear and correct, but understandable.  For example,
if the robot recognizes an object as an {\em orange} but the human
cannot see the object due to view occlusion, a simple referring
expression, such as {\em next to the orange at the corner} will not be
understandable.  Robust models need to consider the human's
perspective and knowledge state. In general, the goal is not to
minimize the speaker's own effort, but rather to minimize the joint
effort to come to a common ground \cite{Clark1996}. Thus a robot
should often make the extra effort to make sure the human understands.  
This may involve, for example, generating a description in small pieces,
giving
the human the chance to give interleaved feedback to verify that the
knowledge states are aligning \cite{fang2015}, or by proactively first
describing essentials of its own internal representation, to make
subsequent grounding more efficient \cite{chai2014}. Future language
generation and speech synthesis modules will need more systematic
techniques for applying theory-of-mind reasoning to model humans'
mental models and perspectives, and methods for collaborative
grounding.
Further, in an environment containing multiple human agents, a robot
needs to design its utterances to make them clear to specific
individuals or groups of individuals, and to craft them to make 
 clear at each time who its utterances are  addressing.

\reccox{S2}{Develop speech generators that support multimodal
  interaction.}

Robots are embodied and multimodal.  To be effective,
actions in the linguistic channel must be coordinated with other
channels, such as physical gestures and eye gaze. This involves not
only selection of appropriate word sequences but also
utterance prosody.  This is especially important for robots that need
to be able to refer to specific objects in the environment, and need
to show ongoing awareness of the environment as things change.

Fortunately, many robots have physical attributes that enable them to
communicate more efficiently. For example, they often have
capabilities for gestures and postures or gaze to show direction of
attention.  Generated language should incorporate deictic expressions
and be coordinated with the timing of a robot's physical gestures.  Robots
with capabilities for facial expressions need to coordinate those with
the timing of prosodic emphasis or intonational cues associated with a
question. In addition, language should be coordinated with path
planning and motion planning, as when a robot needs
to convey that it is about to move {\it over here}.  Robots that have
these abilities will be able to  communicate more efficiently, often
using just a few words deftly augmented with multimodal and prosodic
signals.
Indeed, robot-to-human information transfer may evolve from being
a sequence of individual communicative actions to something more continuous: an ongoing display of state and intention.

\reccox{S3}{Create synthesizers that support realtime control of the  voice.}

Robots operate in real time, so synthesizers must also.  There are
several aspects to this.  As robots must respond to dynamic changes in
the environment, the generation and speaking processes need to be
interruptible and the plans must be modifiable.  For example, human
speakers reflexively pause if a loud noise occurs in the environment,
or if an addressee seems to be not paying attention, and robots should
do the same.  To coordinate spoken language with a robot's physical
gestures and motion, synthesizers must need to be able to output synch
points and to support fine-grained timing control.  Moreover, a
robot's speech may need to be timed to support, guide, or complement
the user's actions and utterances.  Incremental synthesis is also
commonly needed. Timing is thus a major issue, with progress needed on
many fronts.

\reccox{S4}{Develop the ability to tune speech generators to convey a
  desired tone, personality, and identity.}

One can take the perspective that robots should be purely
functional, and that designers should not bother to produce robots
that project a specific personality.  However, designing a robot
to have no detectable personality is itself a
design choice.  It is not uncommon today to hear robots with voices
chosen only on the basis of intelligibility, and this guides the user
to expect a formal, tedious interaction partner.  Many highly capable
agent systems, such as Siri and Alexa, have clear, dominant voices to
convey to the user that they should adopt a formal turn-taking style
and keep their utterances short and to the point.  Yet this strategy
is not always appropriate; for example, a robot interacting with a
small child should talk very differently, and a robot assisting in a
disaster recovery effort should sound different again.

In general, the voice of an artificial agent, together with its visual
appearance, tells the user what to expect of it.  Thus we need voices
that are parameterizable --- to be a little more childlike, more
rigid, more helpless, more businesslike, and so on --- to meet the
needs of an application, and guidelines for making such
choices. Further, although contemporary speech synthesis is capable of
generating utterances virtually indistinguishable from those produced
by a human being, this will often be inappropriate or even unethical:
the use of human-like voices for artificial devices encourages people
to overestimate their linguistic and cognitive capabilities.  In many
cases, the voice should make the agent clearly identifiable as an
artificial individual: a ``robotic'' voice can be natural for a robot;
yet such a voice need not be a low-quality voice, it being perfectly
possible to generate high-quality robotic-sounding speech
\cite{Wilson2017}.
Beyond just the qualities of the voice itself, variation in style more
generally, including word choice and interaction style, can also guide
user expectations, but so far has been very understudied. 

\begin{figure}[t]
 \centering
 \includegraphics[width=5in]{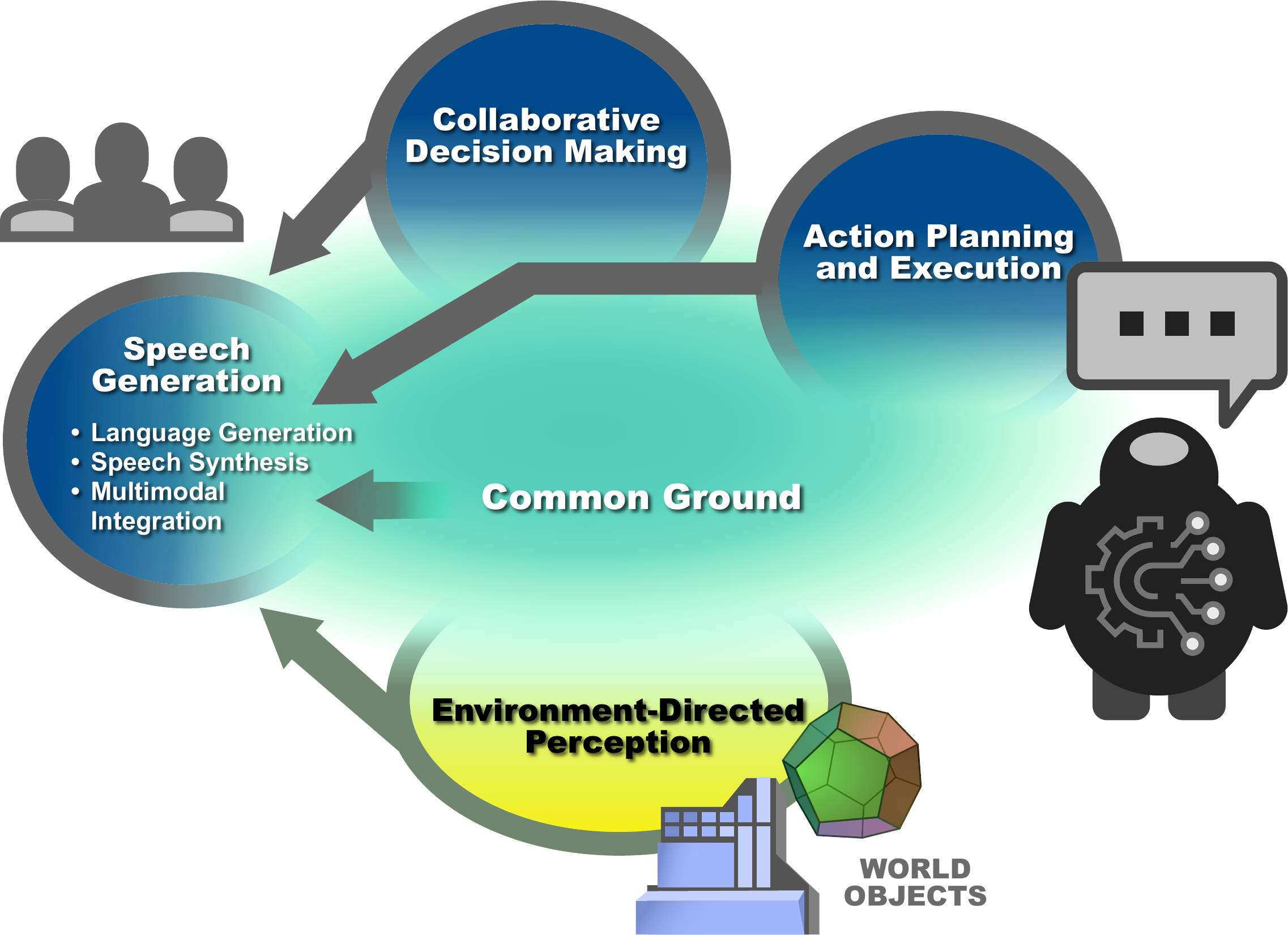}
 \caption{Speech generation in context, derived from Figure \ref{fig:arch}. 
 There the focus was on human-to-robot communication; here, despite the different direction of information flow, many of the same knowledge sources and representations are relevant.}
 \label{fig:synth-figure}
\end{figure}

\medskip
Figure \ref{fig:synth-figure} illustrates some of the points made above.
Speech Generation needs to be to able convey not only propositions and
intents (top arrow), but also other aspects of the robot's internal
state, stance, and knowledge.  Speech Generation also needs to be
appropriately timed and coordinated with robot actions and multimodal
signaling (middle arrow).  To do so effectively, it needs access to
the robot's models of what the user knows, and of the current state of the 
environment (bottom arrows).

%% file: policy.tex
\section{Policy}

Our recommendations to this point have been directed to researchers
and developers: scientific and technical challenges call for
scientific and technical advances.  But to enable these advances, a
healthy research ecosystem is needed, and this also requires  supportive decisions by funders and policy makers. 

\reccox{P1}{Create funding opportunities specifically for spoken language interaction with robots.}

None of the issues noted in this report are entirely new: all have been discussed
before.  Yet they remain as unsolved issues today, in part because
they have tended to fall through the cracks.  Few of the challenges we
have identified are core robotics topics, and few are pure speech
science topics.  In competitions for funding, work on these issues may
seem only marginally appealing to pure robotics programs, and
also marginal to pure speech science programs.  The challenges and
needs in spoken language interaction with robots are specific enough
and non-mainstream enough that funders need to use special care to
nurture this area. While there are bottom-up efforts to
foster research in this area, such as the recent RoboDial workshop, 
top-down support is also critical.

Spoken language interaction with robots is a highly interdisciplinary area,
requiring researchers and developers who are competent in multiple
disciplines --- speech, language, vision, robotics, machine learning,
etc. --- and able to collaborate broadly.  Great benefit could result
from the creation of a multi-disciplinary curriculum or consortium to
train the next generation of researchers and developers. These might
include summer schools or summer camps (perhaps like the human
language technology summer programs organized by Johns Hopkins)
dedicated to the intersection of speech, language, vision, and
robotics.  We also see the need for more fellowship and internship
programs, some perhaps joint public-private efforts, to encourage
Ph.D. students to do multi-disciplinary research that connects spoken
language and robotics.

At the same time, care must be taken to ensure that work in this area
does not become inward-looking, instead remaining well-connected to
other work in speech, language, and robotics.

\reccox{P2}{Prefer evaluation based on use cases.}    

There is an essential tension between intrinsic and extrinsic
evaluation. Today in most areas of speech processing the former is
pervasive: researchers commonly tackle an existing dataset and develop
a new algorithm that improves on previous results according to a
standard metric.  Yet ultimately we need to evaluate research
extrinsically, judged by its contribution towards providing useful
communicative capabilities to users.  Doing so brings greater
likelihood of leading to novel results and perspectives, and of
driving real progress. However, extrinsic evaluation is much
more time-consuming and expensive. This is true especially for interaction, 
as the evaluation
of interactive behaviors cannot really  be done by reference to static data sets.
Even for the best-understood aspects of extrinsic evaluation, relating
to user satisfaction, meaningful measurement is difficult, and the
results depend on so many factors \cite{moeller-engelbrecht2009} that
the generality can always be questioned.  Still, we hope that funders
and proposal reviewers will prefer extrinsic evaluation,
and, at the same time, researchers will work to address the need for
new evaluation methodologies that are both efficient and highly
informative.

\reccox{P3}{Support many kinds of research and development activities. }

Not everything we need to advance this area can be formulated as a
classic research project.  Some things are pure infrastructure
development, including the development of shareable data, software,
and hardware.  Others, like dealing with reverberation, are now
engineering problems more than scientific problems.  We encourage
funding agencies to look at the big picture, and support, when
appropriate, any activity that is critical to progress.  At the same
time, we recognize that federal funding is not the answer to every
question.  The field needs to figure out how industry, including
start-ups, can carve off important problems, solve them, enable those
solutions to be widely used, and make a profit from doing so.

\begin{figure}[t]
    \centering
\includegraphics[width=3in]{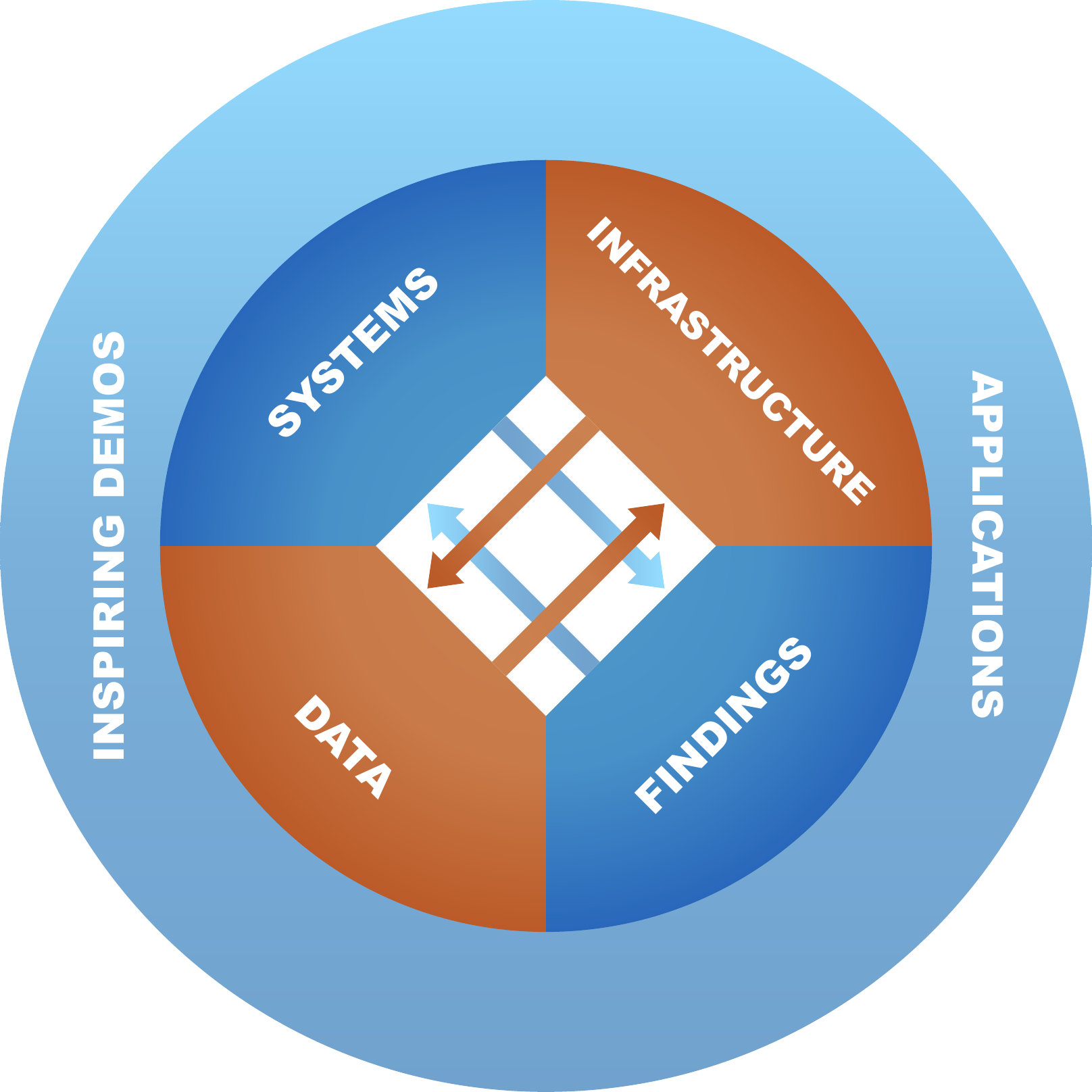}
    \caption{Some of the many facets of speech for robotics. Funders should recognize this diversity of needs and the diversity of methods they require, and be able to support projects that focus on only one aspect or
    approach.}
    \label{fig:fundables}
\end{figure}

One important type of activity to value is truly focused research.
Research solicitations naturally reflect funders' multitudinous
ambitions and desires, and, as a result, they increasingly call for
every research proposal to address all of these at once ---
from the creation of data resources and software infrastructure, to
demo-ing on real robots and addressing true scientific questions, to
doing outreach and technology transfer --- as suggested by Figure
\ref{fig:fundables}.  Yet focused work is also important: advances in
only one or two areas can still be highly significant.  For example, a
research project to answer core questions in this area may not need to
involve real robots.  Similarly, research on
software architectures that make robots' internal representations more
useful for speech-based communication may not need to promise advances
in any core speech or robotics technology.

\reccox{P4}{Work to overcome the barriers to data sharing.}  

Data is the lifeblood of research in spoken language interaction with
robots.  In particular, recordings of real humans interacting with
real or simulated robots can be used for analysis, discovery, and
model training.  Unfortunately, today almost all such data is trapped
within individual institutions, barricaded by restrictions that
prevent sharing.  Many of these restrictions exist for good reasons,
but we still need to work to find ways to share data that are
respectful of privacy concerns.  While we see no simple solutions, we
do see two possible ways to address this problem.  First, researchers
should, whenever possible, design data collections to be fully
shareable, which may in some cases be as simple as having participants
dedicate their ``work'' to the public domain, for example using
the Creative Commons CC0 license.  Second, researchers should try to educate their
local Institutional Review Boards (IRBs) on both the costs and
benefits of tight restrictions on data sharing.  To this end, the
creation of a model IRB proposal that shows how to address the issues
and concerns could be very helpful, especially if endorsed by the
NSF or another prestigious organization.

\reccox{P5}{Explore novel public-private partnerships for open source software.}

Open source software is enormously valuable for both research and
development, but in this area we see a missed opportunity.
Traditionally, open source software comes from universities, but
industry has a great deal of valuable software that might be
released.  While some companies aim mostly to protect intellectual
property, others are potentially willing to open-source some or all of
their products.  Examples include Willow Garage --- and now the Open Source
Robotics Foundation --- which provided Robot Operating System (ROS) free and
open-source, as well as the PR2 as a reference robot hardware system,
both of which catalyzed research in autonomous robotics.  However,
releasing clean, well-documented open-source software takes resources
and is seldom a direct revenue generator, so external encouragement
and support can be necessary.  Thus we see the opportunity for novel
private-public partnership to help create and maintain open reference
implementations of software and hardware for enabling social robotics. We might call this a ``reverse SBIR,''
in the sense that a for-profit company and a research institution
collaborate, not to commercialize a high-risk/high-reward research
project (as with a traditional SBIR), but, rather, to document and
publish open-source software implementations and hardware designs of
company products.  For companies the benefits may include extended
product lifetimes and visibility, and for the research community new
tools and testbeds for research.

\bigskip
Enabling spoken language interaction with robots will require
advances in speech science, in robotics, and at the intersection.
There is a lot to do: not just one single problem to solve, but a
multifaceted challenge, needing attack from many fronts, over years
and decades.  Our hope is that this report, by providing a
clearer view of the key issues, will help researchers and funders
optimally choose what to tackle, and ultimately, after much hard
work, bring us to the day when we can interact with robots
effectively and smoothly, just by talking with them.

%% file: main.bbl
\begin{thebibliography}{117}
\providecommand{\natexlab}[1]{#1}
\providecommand{\url}[1]{\texttt{#1}}
\expandafter\ifx\csname urlstyle\endcsname\relax
  \providecommand{\doi}[1]{doi: #1}\else
  \providecommand{\doi}{doi: \begingroup \urlstyle{rm}\Url}\fi

\bibitem[Admoni and Scassellati(2017)]{admoni2017social}
Henny Admoni and Brian Scassellati.
\newblock Social eye gaze in human-robot interaction: A review.
\newblock \emph{Journal of Human-Robot Interaction}, 6\penalty0 (1):\penalty0
  25--63, 2017.

\bibitem[Al~Moubayed et~al.(2012)Al~Moubayed, Beskow, Skantze, and
  Granstr{\"o}m]{al2012furhat}
Samer Al~Moubayed, Jonas Beskow, Gabriel Skantze, and Bj{\"o}rn Granstr{\"o}m.
\newblock Furhat: A back-projected human-like robot head for multiparty
  human-machine interaction.
\newblock In \emph{Cognitive Behavioural Systems}, pages 114--130. Springer,
  2012.

\bibitem[Amershi et~al.(2019)Amershi, Weld, Vorvoreanu, Fourney, Nushi,
  Collisson, Suh, Iqbal, Bennett, Inkpen, Teevan, Kikin-Gil, and
  Horvitz]{amershi2019guidelines}
Saleema Amershi, Dan Weld, Mihaela Vorvoreanu, Adam Fourney, Besmira Nushi,
  Penny Collisson, Jina Suh, Shamsi Iqbal, Paul~N. Bennett, Kori Inkpen, Jaime
  Teevan, Ruth Kikin-Gil, and Eric Horvitz.
\newblock Guidelines for human-{AI} interaction.
\newblock In \emph{Proceedings of CHI}, pages 1--13, 2019.

\bibitem[Anderson et~al.(2018)Anderson, Wu, Teney, Bruce, Johnson,
  S{\"u}nderhauf, Reid, Gould, and van~den Hengel]{anderson2018}
Peter Anderson, Qi~Wu, Damien Teney, Jake Bruce, Mark Johnson, Niko
  S{\"u}nderhauf, Ian Reid, Stephen Gould, and Anton van~den Hengel.
\newblock Vision-and-{{Language Navigation}}: {{Interpreting}}
  visually-grounded navigation instructions in real environments.
\newblock In \emph{Proceedings of CVPR}, pages 3674--3683, 2018.

\bibitem[Argall et~al.(2009)Argall, Chernova, Veloso, and
  Browning]{argall2009survey}
Brenna~D. Argall, Sonia Chernova, Manuela Veloso, and Brett Browning.
\newblock A survey of robot learning from demonstration.
\newblock \emph{Robotics and Autonomous Systems}, 57\penalty0 (5):\penalty0
  469--483, 2009.

\bibitem[Artzi and Zettlemoyer(2013)]{artzi13a}
Yoav Artzi and Luke Zettlemoyer.
\newblock Weakly supervised learning of semantic parsers for mapping
  instructions to actions.
\newblock \emph{Transactions of the Association for Computational Linguistics},
  1:\penalty0 49--62, 2013.

\bibitem[Bainbridge et~al.(2011)Bainbridge, Hart, Kim, and
  Scassellati]{bainbridge11}
Wilma~A. Bainbridge, Justin~W. Hart, Elizabeth~S. Kim, and Brian Scassellati.
\newblock The benefits of interactions with physically present robots over
  video-displayed agents.
\newblock \emph{International Journal of Social Robotics}, 3\penalty0
  (1):\penalty0 41--52, 2011.

\bibitem[Balentine(2007)]{Balentine2007}
Bruce Balentine.
\newblock \emph{{It's Better to Be a Good Machine Than a Bad Person: Speech
  Recognition and Other Exotic User Interfaces at the Twilight of the Jetsonian
  Age}}.
\newblock ICMI Press, 2007.

\bibitem[Bangerter and Clark(2003)]{bangerter2003navigating}
Adrian Bangerter and Herbert~H. Clark.
\newblock Navigating joint projects with dialogue.
\newblock \emph{Cognitive Science}, 27\penalty0 (2):\penalty0 195--225, 2003.

\bibitem[Baumann and Schlangen(2012)]{Baumann2012}
Timo Baumann and David Schlangen.
\newblock {The InproTK 2012 Release}.
\newblock In \emph{Proceedings of the NAACL-HLT Workshop on Future Directions
  and Needs in the Spoken Dialog Community: Tools and Data}, pages 29--32,
  2012.

\bibitem[Baumann et~al.(2017)Baumann, Kennington, Hough, and
  Schlangen]{Baumann2017}
Timo Baumann, Casey Kennington, Julian Hough, and David Schlangen.
\newblock {Recognising conversational speech: What an incremental ASR should do
  for a dialogue system and how to get there}.
\newblock In Kristiina Jokinen and Graham Wilcock, editors, \emph{Dialogues
  with Social Robots}, pages 421--432. Springer, 2017.

\bibitem[Beckerle et~al.(2019)Beckerle, Castellini, and
  Lenggenhager]{beckerle2019robotic}
Philipp Beckerle, Claudio Castellini, and Bigna Lenggenhager.
\newblock Robotic interfaces for cognitive psychology and embodiment research:
  {A} research roadmap.
\newblock \emph{Wiley Interdisciplinary Reviews: Cognitive Science},
  10\penalty0 (2), 2019.

\bibitem[Blukis et~al.(2019)Blukis, Terme, Niklasson, Knepper, and
  Artzi]{blukis19a}
Valts Blukis, Yannick Terme, Eyvind Niklasson, Ross~A. Knepper, and Yoav Artzi.
\newblock Learning to map natural language instructions to physical quadcopter
  control using simulated flight.
\newblock In \emph{Proceedings of the Conference on Robot Learning}, pages
  1415--1438, 2019.

\bibitem[Bohus and Horvitz(2009)]{bohus2009models}
Dan Bohus and Eric Horvitz.
\newblock Models for multiparty engagement in open-world dialog.
\newblock In \emph{Proceedings of SIGdial}, pages 225--234, 2009.

\bibitem[Bohus and Horvitz(2010)]{bohus2010challenges}
Dan Bohus and Eric Horvitz.
\newblock On the challenges and opportunities of physically situated dialog.
\newblock In \emph{Proceedings of the AAAI Fall Symposium on Dialog with
  Robots}, 2010.

\bibitem[Bohus et~al.(2014)Bohus, Saw, and Horvitz]{bohus2014directions}
Dan Bohus, Chit~W. Saw, and Eric Horvitz.
\newblock Directions robot: In-the-wild experiences and lessons learned.
\newblock In \emph{Proceedings of AAMAS}, pages 637--644, 2014.

\bibitem[Bohus et~al.(2017)Bohus, Andrist, and Jalobeanu]{Bohus2017}
Dan Bohus, Sean Andrist, and Mihai Jalobeanu.
\newblock {Rapid Development of Multimodal Interactive Systems: A Demonstration
  of Platform for Situated Intelligence}.
\newblock In \emph{Proceedings of ICMI}, pages 493--494, 2017.

\bibitem[Boltz(2005)]{boltz05}
Marilyn~G. Boltz.
\newblock Temporal dimensions of conversational interaction: The role of
  response latencies and pauses in social impression formation.
\newblock \emph{Journal of Language and Social Psychology}, 24\penalty0
  (2):\penalty0 103--138, 2005.

\bibitem[Boteanu et~al.(2016)Boteanu, Howard, Arkin, and
  Kress-Gazit]{boteanu2016verifiable}
Adrian Boteanu, Thomas~M. Howard, Jacob Arkin, and Hadas Kress-Gazit.
\newblock A model for verifiable grounding and execution of complex natural
  language instructions.
\newblock In \emph{Proceedings of IROS}, pages 2649--2654, 2016.

\bibitem[Boteanu et~al.(2017)Boteanu, Arkin, Patki, Howard, and
  Kress-Gazit]{boteanu2017robotinitiatedinteraction}
Adrian Boteanu, Jacob Arkin, Siddharth Patki, Thomas~M. Howard, and Hadas
  Kress-Gazit.
\newblock Robot-initiated specification repair through grounded language
  interaction.
\newblock In \emph{Proceedings of the AAAI Fall Symposium on Natural
  Communication for Human-Robot Collaboration}, 2017.

\bibitem[Broekens et~al.(2009)Broekens, Heerink, and
  Rosendal]{broekens2009assistive}
Joost Broekens, Marcel Heerink, and Henk Rosendal.
\newblock Assistive social robots in elderly care: a review.
\newblock \emph{Gerontechnology}, 8\penalty0 (2):\penalty0 94--103, 2009.

\bibitem[Bulyko and Ostendorf(2001)]{bulyko2001}
Ivan Bulyko and Mari Ostendorf.
\newblock Joint prosody prediction and unit selection for concatenative speech
  synthesis.
\newblock In \emph{Proceedings of ICASSP}, volume~2, pages 781--784, 2001.

\bibitem[Bu{\ss} and Schlangen(2011)]{2300868}
Okko Bu{\ss} and David Schlangen.
\newblock {DIUM -- An incremental dialogue manager that can produce
  self-corrections}.
\newblock In \emph{Proceedings of SemDial}, 2011.

\bibitem[Chai et~al.(2014)Chai, She, Fang, Ottarson, Littley, Liu, and
  Hanson]{chai2014}
Joyce~Y. Chai, Lanbo She, Rui Fang, Spencer Ottarson, Cody Littley, Changsong
  Liu, and Kenneth Hanson.
\newblock Collaborative effort towards common ground in situated human-robot
  dialogue.
\newblock In \emph{Proceedings of HRI}, pages 33--40, 2014.

\bibitem[Chai et~al.(2018)Chai, Gao, She, Yang, {Saba-Sadiya}, and
  Xu]{chai2018}
Joyce~Y. Chai, Qiaozi Gao, Lanbo She, Shaohua Yang, Sari {Saba-Sadiya}, and
  Guangyue Xu.
\newblock Language to {Action}: {Towards Interactive Task Learning} with
  {Physical Agents}.
\newblock In \emph{Proceedings of IJCAI}, pages 2--9, 2018.

\bibitem[Chang et~al.(2019)Chang, Hauptmann, Morency, Antani, Bulterman, Busso,
  Chai, Hirschberg, Jain, Mayer-Patel, Meth, Mooney, Nahrstedt, Narayanan,
  Natarajan, Oviatt, Prabhakaran, Smeulders, Sundaram, Zhang, and
  Zhou]{nsf-multimedia}
Shih-Fu Chang, Alex Hauptmann, Louis-Philippe Morency, Sameer Antani, Dick
  Bulterman, Carlos Busso, Joyce Chai, Julia Hirschberg, Ramesh Jain, Ketan
  Mayer-Patel, Reuven Meth, Raymond Mooney, Klara Nahrstedt, Shri Narayanan,
  Prem Natarajan, Sharon Oviatt, Balakrishnan Prabhakaran, Arnold Smeulders,
  Hari Sundaram, Zhengyou Zhang, and Michelle Zhou.
\newblock Report of 2017 {NSF Workshop on Multimedia Challenges, Opportunities
  and Research Roadmaps}.
\newblock arXiv preprint arXiv:1908.02308, 2019.

\bibitem[Chen and Mooney(2011)]{chen2011learning}
David~L. Chen and Raymond~J. Mooney.
\newblock Learning to interpret natural language navigation instructions from
  observations.
\newblock In \emph{Proceedings of AAAI}, pages 859--865, 2011.

\bibitem[Chen and Mostow(2011)]{chen2011tale}
Wei Chen and Jack Mostow.
\newblock A tale of two tasks: Detecting children's off-task speech in a
  reading tutor.
\newblock In \emph{Proceedings of Interspeech}, pages 1621--1624, 2011.

\bibitem[Chhetri et~al.(2018)Chhetri, Hilmes, Kristjansson, Chu, Mansour, Li,
  and Zhang]{Chhetri2018}
Amit Chhetri, Philip Hilmes, Trausti Kristjansson, Wai Chu, Mohamed Mansour,
  Xiaoxue Li, and Xianxian Zhang.
\newblock Multichannel audio front-end for far-field automatic speech
  recognition.
\newblock In \emph{Proceedings of the 26th European Signal Processing
  Conference (EUSIPCO)}, pages 1527--1531, 2018.

\bibitem[Christensen et~al.(2009)Christensen, Batzinger, Bekris, Bohringer,
  Bordogna, Bradski, Brock, Burnstein, Fuhlbrigge, Eastman, Edsinger, Fuchs,
  Goldberg, Henderson, Joyner, Kavaraki, Kelly, Kelly, Kumar, Manocha,
  McCallum, Mosterman, Messina, Murphey, Peters, Shepherd, Singh, Sweet,
  Trinkle, Tsai, Wells, Wurman, Yorio, and Zhang]{robotics-roadmap}
Henrik~I. Christensen, Tom Batzinger, Kostas Bekris, Karl Bohringer, Joe
  Bordogna, Gary Bradski, Oliver Brock, Jeff Burnstein, Thomas Fuhlbrigge,
  Roger Eastman, Aaron Edsinger, Erica Fuchs, Ken Goldberg, Tom Henderson,
  William Joyner, Lydia Kavaraki, Clint Kelly, Alonzo Kelly, Vijay Kumar,
  Dinesh Manocha, Andrew McCallum, Pieter Mosterman, Elena Messina, Todd
  Murphey, Richard~Alan Peters, Stuart Shepherd, Sanjiv Singh, Larry Sweet,
  Jeff Trinkle, Jason Tsai, James Wells, Peter Wurman, Tom Yorio, and Mingjun
  Zhang.
\newblock \emph{A Roadmap for {US} Robotics: From Internet to Robotics}.
\newblock Computing Community Consortium (CCC), 2009.

\bibitem[Clark(1996)]{Clark1996}
Herbert~H. Clark.
\newblock \emph{Using Language}.
\newblock Cambridge University Press, 1996.

\bibitem[Cohen et~al.(2004)Cohen, Giangola, and Balogh]{dog-book}
Michael~H. Cohen, James~P. Giangola, and Jennifer Balogh.
\newblock \emph{Voice User Interface Design}.
\newblock Addison-Wesley, 2004.

\bibitem[Cummins et~al.(2015)Cummins, Scherer, Krajewski, Schneider, Epps, and
  Quatieri]{cummins2015}
Nicholas Cummins, Stefan~C. Scherer, Jarek Krajewski, Sebastian Schneider,
  Julien Epps, and Thomas~F. Quatieri.
\newblock A review of depression and suicide risk assessment using speech
  analysis.
\newblock \emph{Speech Communication}, 71:\penalty0 10--49, 2015.

\bibitem[Deng et~al.(2019)Deng, Mutlu, and Matari{\'c}]{deng2019embodiment}
Eric Deng, Bilge Mutlu, and Maja~J. Matari{\'c}.
\newblock Embodiment in socially interactive robots.
\newblock \emph{Foundations and Trends® in Robotics}, 7\penalty0 (4):\penalty0
  251--356, 2019.

\bibitem[{DeVault} et~al.(2014){DeVault}, Artstein, Benn, Dey, Fast, Gainer,
  Georgila, Gratch, Hartholt, Lhommet, Lucas, Marsella, Morbini, Nazarian,
  Scherer, Stratou, Suri, Traum, Wood, Xu, Rizzo, and Morency]{devault-aamas14}
David {DeVault}, Ron Artstein, Grace Benn, Teresa Dey, Ed~Fast, Alesia Gainer,
  Kallirroi Georgila, Jon Gratch, Arno Hartholt, Margaux Lhommet, Gale Lucas,
  Stacy Marsella, Fabrizio Morbini, Angela Nazarian, Stefan Scherer, Giota
  Stratou, Apar Suri, David Traum, Rachel Wood, Yuyu Xu, Albert Rizzo, and
  Louis-Philippe Morency.
\newblock {SimSensei Kiosk}: A virtual human interviewer for healthcare
  decision support.
\newblock In \emph{Proceedings of AAMAS}, pages 1061--1068, 2014.

\bibitem[Devillers et~al.(2020)Devillers, Kawahara, Moore, and
  Scheutz]{slivar2020}
Laurence Devillers, Tatsuya Kawahara, Roger~K. Moore, and Matthias Scheutz.
\newblock Spoken language interaction with virtual agents and robots
  ({SLIVAR}): Towards effective and ethical interaction ({D}agstuhl {S}eminar
  20021).
\newblock In \emph{Dagstuhl Reports}, volume~10. Schloss
  Dagstuhl-Leibniz-Zentrum f{\"u}r Informatik, 2020.

\bibitem[El~Shafey et~al.(2019)El~Shafey, Soltau, and Shafran]{shafey2019}
Laurent El~Shafey, Hagen Soltau, and Izhak Shafran.
\newblock Joint speech recognition and speaker diarization via sequence
  transduction.
\newblock In \emph{Proceedings of Interspeech}, pages 396--400, 2019.

\bibitem[Eskenazi and Zhao(2020)]{eskenazi2020report}
Maxine Eskenazi and Tiancheng Zhao.
\newblock Report from the {NSF Future Directions Workshop}, {T}oward
  user-oriented agents: Research directions and challenges.
\newblock arXiv preprint arXiv:2006.06026, 2020.

\bibitem[Fang et~al.(2015)Fang, Doering, and Chai]{fang2015}
Rui Fang, Malcolm Doering, and Joyce~Y. Chai.
\newblock Embodied collaborative referring expression generation in situated
  human-robot dialogue.
\newblock In \emph{Proceedings of HRI}, pages 271--278, 2015.

\bibitem[Fong et~al.(2003)Fong, Thorpe, and Baur]{fong2003collaboration}
Terrence Fong, Charles Thorpe, and Charles Baur.
\newblock Collaboration, dialogue, and human-robot interaction.
\newblock In \emph{Robotics Research}, pages 255--266. Springer, 2003.

\bibitem[Forbes-Riley and Litman(2011)]{litman-uncertainty11b}
Kate Forbes-Riley and Diane Litman.
\newblock Benefits and challenges of real-time uncertainty detection and
  adaptation in a spoken dialogue computer tutor.
\newblock \emph{Speech Communication}, 53\penalty0 (9-10):\penalty0 1115--1136,
  2011.

\bibitem[Fusaroli and Tyl{\'e}n(2016)]{fusaroli16}
Riccardo Fusaroli and Kristian Tyl{\'e}n.
\newblock Investigating conversational dynamics: Interactive alignment,
  interpersonal synergy, and collective task performance.
\newblock \emph{Cognitive Science}, 40\penalty0 (1):\penalty0 145--171, 2016.

\bibitem[Gaschler et~al.(2012)Gaschler, Jentzsch, Giuliani, Huth, de~Ruiter,
  and Knoll]{gaschler2012social}
Andre Gaschler, S{\"o}ren Jentzsch, Manuel Giuliani, Kerstin Huth, Jan
  de~Ruiter, and Alois Knoll.
\newblock Social behavior recognition using body posture and head pose for
  human-robot interaction.
\newblock In \emph{Proceedings of IROS}, pages 2128--2133, 2012.

\bibitem[Gatt and Krahmer(2018)]{gatt-krahmer}
Albert Gatt and Emiel Krahmer.
\newblock Survey of the state of the art in natural language generation: Core
  tasks, applications and evaluation.
\newblock \emph{Journal of Artificial Intelligence Research}, 61:\penalty0
  65--170, 2018.

\bibitem[Geertzen(2015)]{geertzen15}
Jeroen Geertzen.
\newblock Exploring age-related conversational interaction.
\newblock In \emph{Proceedings of SemDial}, pages 42--47, 2015.

\bibitem[Giannakopoulos(2015)]{giannakopoulos2015pyaudioanalysis}
Theodoros Giannakopoulos.
\newblock {pyAudioAnalysis}: An open-source python library for audio signal
  analysis.
\newblock \emph{PloS one}, 10\penalty0 (12), 2015.

\bibitem[Gil and Selman(2019)]{ai-roadmap19}
Yolanda Gil and Bart Selman.
\newblock \emph{A 20-Year Community Roadmap for Artificial Intelligence
  Research in the US}.
\newblock Computing Community Consortium (CCC) and Association for the
  Advancement of Artificial Intelligence (AAAI), 2019.

\bibitem[Giles et~al.(1987)Giles, Mulac, Bradac, and Johnson]{giles-mulac}
Howard Giles, Anthony Mulac, James~J. Bradac, and Patricia Johnson.
\newblock Speech accommodation theory: The first decade and beyond.
\newblock In \emph{Annals of the International Communication Association},
  volume~10, pages 13--48. Taylor \& Francis, 1987.

\bibitem[Glas et~al.(2016)Glas, Minato, Ishi, Kawahara, and
  Ishiguro]{glas2016erica}
Dylan~F. Glas, Takashi Minato, Carlos~T. Ishi, Tatsuya Kawahara, and Hiroshi
  Ishiguro.
\newblock Erica: The {ERATO} intelligent conversational android.
\newblock In \emph{Proceedings of RO-MAN}, pages 22--29, 2016.

\bibitem[Gockley and Matari{\'c}(2006)]{gockley2006encouraging}
Rachel Gockley and Maja~J Matari{\'c}.
\newblock Encouraging physical therapy compliance with a hands-off mobile
  robot.
\newblock In \emph{Proceedings of HRI}, pages 150--155, 2006.

\bibitem[Gratch et~al.(2007)Gratch, Wang, Okhmatovskaia, Lamothe, Morales,
  van~der Werf, and Morency]{gratch-engaging}
Jonathan Gratch, Ning Wang, Anna Okhmatovskaia, Francois Lamothe, Mathieu
  Morales, Rick~J. van~der Werf, and Louis-Philippe Morency.
\newblock Can virtual humans be more engaging than real ones?
\newblock In \emph{Proceedings of HCI}, pages 286--297, 2007.

\bibitem[Grothendieck et~al.(2011)Grothendieck, Gorin, and
  Borges]{grothendieck11}
John Grothendieck, Allen~L. Gorin, and Nash Borges.
\newblock Social correlates of turn-taking style.
\newblock \emph{Computer Speech and Language}, 25\penalty0 (4):\penalty0
  789--801, 2011.

\bibitem[Hirst et~al.(1994)Hirst, McRoy, Heeman, Edmonds, and
  Horton]{hirst1994repairing}
Graeme Hirst, Susan McRoy, Peter Heeman, Philip Edmonds, and Diane Horton.
\newblock Repairing conversational misunderstandings and non-understandings.
\newblock \emph{Speech Communication}, 15\penalty0 (3-4):\penalty0 213--229,
  1994.

\bibitem[Hoegen et~al.(2019)Hoegen, Aneja, McDuff, and Czerwinski]{hoegen19}
Rens Hoegen, Deepali Aneja, Daniel McDuff, and Mary Czerwinski.
\newblock An end-to-end conversational style matching agent.
\newblock In \emph{Proceedings of IVA}, pages 111--118, 2019.

\bibitem[Hough et~al.(2015)Hough, Kennington, Schlangen, and
  Ginzburg]{Hough2015}
Julian Hough, Casey Kennington, David Schlangen, and Jonathan Ginzburg.
\newblock Incremental semantics for dialogue processing: Requirements, and a
  comparison of two approaches.
\newblock In \emph{Proceedings of the International Conference on Computational
  Semantics}, pages 206--216, 2015.

\bibitem[Howard et~al.(2014)Howard, Tellex, and Roy]{howard2014natural}
Thomas~M. Howard, Stefanie Tellex, and Nicholas Roy.
\newblock A natural language planner interface for mobile manipulators.
\newblock In \emph{Proceedings of ICRA}, pages 6652--6659, 2014.

\bibitem[Hudry et~al.(2013)Hudry, Aldred, Wigham, Green, Leadbitter, Temple,
  Barlow, and McConachie]{hudry13}
Kristelle Hudry, Catherine Aldred, Sarah Wigham, Jonathan Green, Kathy
  Leadbitter, Kathryn Temple, Katherine Barlow, and Helen McConachie.
\newblock Predictors of parent--child interaction style in dyads with autism.
\newblock \emph{Research in Developmental Disabilities}, 34\penalty0
  (10):\penalty0 3400--3410, 2013.

\bibitem[Isik et~al.(2016)Isik, Le~Roux, Chen, Watanabe, and Hershey]{isik2016}
Yusuf Isik, Jonathan Le~Roux, Zhuo Chen, Shinji Watanabe, and John~R. Hershey.
\newblock Single-channel multi-speaker separation using deep clustering.
\newblock In \emph{Proceedings of Interspeech}, pages 545--549, 2016.

\bibitem[Itoh et~al.(2009)Itoh, Kitaoka, and Nishimura]{kitaoka09}
Toshihiko Itoh, Norihide Kitaoka, and Ryota Nishimura.
\newblock Subject experiments on influence of response timing in spoken
  dialogues.
\newblock In \emph{Proceedings of Interspeech}, pages 1835--1838, 2009.

\bibitem[Kennington and Schlangen(2017)]{Kennington2017a}
Casey Kennington and David Schlangen.
\newblock A simple generative model of incremental reference resolution for
  situated dialogue.
\newblock \emph{Computer Speech and Language}, 41:\penalty0 43--67, 2017.

\bibitem[Kennington et~al.(2020)Kennington, Moro, Marchand, Carns, and
  McNeill]{Kennington2020}
Casey Kennington, Daniele Moro, Lucas Marchand, Jake Carns, and David McNeill.
\newblock {rrSDS: Towards a Robot-ready Spoken Dialogue System}.
\newblock In \emph{Proceedings of SIGdial}, pages 132--135, 2020.

\bibitem[Kousidis et~al.(2014)Kousidis, Kennington, Baumann, Buschmeier, Kopp,
  and Schlangen]{kousidis2014multimodal}
Spyros Kousidis, Casey Kennington, Timo Baumann, Hendrik Buschmeier, Stefan
  Kopp, and David Schlangen.
\newblock A multimodal in-car dialogue system that tracks the driver's
  attention.
\newblock In \emph{Proceedings of ICMI}, pages 26--33, 2014.

\bibitem[Kruijff et~al.(2007)Kruijff, Zender, Jensfelt, and
  Christensen]{kruijff2007situated}
Geert-Jan~M. Kruijff, Hendrik Zender, Patric Jensfelt, and Henrik~I.
  Christensen.
\newblock Situated dialogue and spatial organization: What, where... and why?
\newblock \emph{International Journal of Advanced Robotic Systems}, 4\penalty0
  (1):\penalty0 125--138, 2007.

\bibitem[Kruijff-Korbayov{\'a} et~al.(2015)Kruijff-Korbayov{\'a}, Colas,
  Gianni, Pirri, de~Greeff, Hindriks, Neerincx, {\"O}gren, Svoboda, and
  Worst]{kruijff2015tradr}
Ivana Kruijff-Korbayov{\'a}, Francis Colas, Mario Gianni, Fiora Pirri, Joachim
  de~Greeff, Koen Hindriks, Mark Neerincx, Petter {\"O}gren, Tom{\'a}{\v{s}}
  Svoboda, and Rainer Worst.
\newblock {TRADR} project: Long-term human-robot teaming for robot assisted
  disaster response.
\newblock \emph{KI-K{\"u}nstliche Intelligenz}, 29\penalty0 (2):\penalty0
  193--201, 2015.

\bibitem[Lazarte(2017)]{iso-robots-rescue}
Maria Lazarte.
\newblock Robots to the rescue!
\newblock \url{https://www.iso.org/news/Ref2169.htm}, 2017.
\newblock Accessed 2020-10-29.

\bibitem[Levitan et~al.(2012)Levitan, Gravano, Willson, Benus, Hirschberg, and
  Nenkova]{levitan12}
Rivka Levitan, Agustin Gravano, Laura Willson, Stefan Benus, Julia Hirschberg,
  and Ani Nenkova.
\newblock Acoustic-prosodic entrainment and social behavior.
\newblock In \emph{Proceedings of NAACL-HLT}, pages 11--19, 2012.

\bibitem[Liu and Chai(2015)]{liu2015aaai}
Changsong Liu and Joyce~Y. Chai.
\newblock Learning to mediate perceptual differences in situated human-robot
  dialogue.
\newblock In \emph{Proceedings of AAAI}, pages 2288--2294, 2015.

\bibitem[Luger and Sellen(2016)]{luger2016like}
Ewa Luger and Abigail Sellen.
\newblock ``{L}ike having a really bad {PA}'': The gulf between user
  expectation and experience of conversational agents.
\newblock In \emph{Proceedings of CHI}, pages 5286--5297, 2016.

\bibitem[Luo and Mesgarani(2019)]{mesgarani2019}
Yi~Luo and Nima Mesgarani.
\newblock {Conv-TasNet}: Surpassing ideal time–frequency magnitude masking
  for speech separation.
\newblock \emph{IEEE/ACM Transactions on Audio, Speech, and Language
  Processing}, 27\penalty0 (8):\penalty0 1256--1266, 2019.

\bibitem[MacMahon et~al.(2006)MacMahon, Stankiewicz, and Kuipers]{macmahon06a}
Matt MacMahon, Brian Stankiewicz, and Benjamin Kuipers.
\newblock Walk the talk: Connecting language, knowledge, and action in route
  instructions.
\newblock In \emph{Proceedings of AAAI}, pages 1475--1482, 2006.

\bibitem[Marge and Rudnicky(2019)]{marge19-tiis}
Matthew Marge and Alexander~I. Rudnicky.
\newblock Miscommunication detection and recovery in situated human–robot
  dialogue.
\newblock \emph{ACM Transactions on Interactive Intelligent Systems},
  9\penalty0 (1), 2019.

\bibitem[Marge et~al.(2019)Marge, Nogar, Hayes, Lukin, Bloecker, Holder, and
  Voss]{marge2019research}
Matthew Marge, Stephen Nogar, Cory~J. Hayes, Stephanie~M. Lukin, Jesse
  Bloecker, Eric Holder, and Clare Voss.
\newblock A {R}esearch {P}latform for {M}ulti-{R}obot {D}ialogue with {H}umans.
\newblock In \emph{Proceedings of NAACL (Demonstrations)}, pages 132--137,
  2019.

\bibitem[Marge et~al.(2020)Marge, Gervits, Briggs, Scheutz, and
  Roque]{marge-etal-2020-first}
Matthew Marge, Felix Gervits, Gordon Briggs, Matthias Scheutz, and Antonio
  Roque.
\newblock Let's do that first! {A} comparative analysis of instruction-giving
  in human-human and human-robot situated dialogue.
\newblock In \emph{Proceedings of SemDial}, 2020.

\bibitem[Matsuyama et~al.(2015)Matsuyama, Akiba, Fujie, and
  Kobayashi]{matsuyama2015four}
Yoichi Matsuyama, Iwao Akiba, Shinya Fujie, and Tetsunori Kobayashi.
\newblock Four-participant group conversation: A facilitation robot controlling
  engagement density as the fourth participant.
\newblock \emph{Computer Speech and Language}, 33\penalty0 (1):\penalty0 1--24,
  2015.

\bibitem[Matuszek et~al.(2013)Matuszek, Herbst, Zettlemoyer, and
  Fox]{matuszek2013learning}
Cynthia Matuszek, Evan Herbst, Luke Zettlemoyer, and Dieter Fox.
\newblock Learning to parse natural language commands to a robot control
  system.
\newblock In \emph{Proceedings of ISER}, pages 403--415, 2013.

\bibitem[Mei et~al.(2016)Mei, Bansal, and Walter]{mei2016listen}
Hongyuan Mei, Mohit Bansal, and Matthew~R. Walter.
\newblock Listen, attend, and walk: Neural mapping of navigational instructions
  to action sequences.
\newblock In \emph{Proceedings of AAAI}, pages 2772--2778, 2016.

\bibitem[Metcalf et~al.(2019)Metcalf, Theobald, Weinberg, Lee, Jonsson, Webb,
  and Apostoloff]{metcalf2019mirroring}
Katherine Metcalf, Barry-John Theobald, Garrett Weinberg, Robert Lee, Ing-Marie
  Jonsson, Russ Webb, and Nicholas Apostoloff.
\newblock Mirroring to build trust in digital assistants.
\newblock In \emph{Proceedings of Interspeech}, pages 4000--4004, 2019.

\bibitem[Michael(2020)]{thilomichael2019a}
Thilo Michael.
\newblock {RETICO}: An incremental framework for spoken dialogue systems.
\newblock In \emph{Proceedings of SIGdial}, pages 49--52, 2020.

\bibitem[Misra et~al.(2017)Misra, Langford, and Artzi]{misra17}
Dipendra Misra, John Langford, and Yoav Artzi.
\newblock Mapping instructions and visual observations to actions with
  reinforcement learning.
\newblock \emph{Proceedings of EMNLP}, pages 1004--1015, 2017.

\bibitem[Moller et~al.(2009)Moller, Engelbrecht, Kuhnel, Wechsung, and
  Weiss]{moeller-engelbrecht2009}
Sebastian Moller, Klaus-Peter Engelbrecht, Christine Kuhnel, Ina Wechsung, and
  Benjamin Weiss.
\newblock A taxonomy of quality of service and quality of experience of
  multimodal human-machine interaction.
\newblock In \emph{Proceedings of the International Workshop on Quality of
  Multimedia Experience}, pages 7--12, 2009.

\bibitem[Moore(2015)]{moore2015talking}
Roger~K. Moore.
\newblock From talking and listening robots to intelligent communicative
  machines.
\newblock In Judith Markowitz, editor, \emph{Robots That Talk and Listen},
  pages 317--335. de Gruyter, 2015.

\bibitem[Moore(2017{\natexlab{a}})]{Moore2017}
Roger~K. Moore.
\newblock {Appropriate voices for artefacts: Some key insights}.
\newblock In \emph{Proceedings of the 1st International Workshop on Vocal
  Interactivity in-and-between Humans, Animals and Robots (VIHAR)}, pages
  7--11, 2017{\natexlab{a}}.

\bibitem[Moore(2017{\natexlab{b}})]{moore2017spoken}
Roger~K. Moore.
\newblock Is spoken language all-or-nothing? {I}mplications for future
  speech-based human-machine interaction.
\newblock In Kristiina Jokinen and Graham Wilcock, editors, \emph{Dialogues
  with Social Robots}, pages 281--291. Springer, 2017{\natexlab{b}}.

\bibitem[Mori(1970)]{Mori1970}
Masahiro Mori.
\newblock {Bukimi no tani (The uncanny valley)}.
\newblock \emph{Energy}, 7:\penalty0 33--35, 1970.

\bibitem[{National Science Foundation}(2020)]{nsf-robotics}
{National Science Foundation}.
\newblock Foundational research in robotics ({S}olicitation).
\newblock \url{https://www.nsf.gov/funding/pgm\_summ.jsp?pims\_id=505784},
  2020.
\newblock Accessed 2020-09-23.

\bibitem[Oviatt et~al.(1997)Oviatt, DeAngeli, and Kuhn]{oviatt1997integration}
Sharon Oviatt, Antonella DeAngeli, and Karen Kuhn.
\newblock Integration and synchronization of input modes during multimodal
  human-computer interaction.
\newblock In \emph{Proceedings of CHI}, pages 415--422, 1997.

\bibitem[Park et~al.(2019)Park, Jia, Bansal, and Manocha]{park2019efficient}
Jae~Sung Park, Biao Jia, Mohit Bansal, and Dinesh Manocha.
\newblock Efficient generation of motion plans from attribute-based natural
  language instructions using dynamic constraint mapping.
\newblock In \emph{Procedings of ICRA}, pages 6964--6971, 2019.

\bibitem[Paul et~al.(2017)Paul, Arkin, Roy, and Howard]{paul2017grounding}
Rohan Paul, Jacob Arkin, Nicholas Roy, and Thomas~M. Howard.
\newblock Grounding abstract spatial concepts for language interaction with
  robots.
\newblock In \emph{Proceedings of IJCAI}, pages 4929--4933, 2017.

\bibitem[Paul et~al.(2018)Paul, Arkin, Aksaray, Roy, and Howard]{paul18a}
Rohan Paul, Jacob Arkin, Derya Aksaray, Nicholas Roy, and Thomas~M. Howard.
\newblock Efficient grounding of abstract spatial concepts for natural language
  interaction with robot platforms.
\newblock \emph{International Journal of Robotics Research}, 37\penalty0
  (10):\penalty0 1269--1299, 2018.

\bibitem[Phillips(2006)]{Philips2006}
Mike Phillips.
\newblock Applications of spoken language technology and systems.
\newblock In \emph{Proceedings of the IEEE/ACL Workshop on Spoken Language
  Technology (SLT)}, page~7. IEEE, 2006.

\bibitem[Ramachandran et~al.(2019)Ramachandran, Sebo, and
  Scassellati]{ramachandran19}
Aditi Ramachandran, Sarah~Strohkorb Sebo, and Brian Scassellati.
\newblock Personalized {R}obot {T}utoring using the {A}ssistive {T}utor {POMDP}
  ({AT-POMDP}).
\newblock In \emph{Proceedings of AAAI}, volume~33, pages 8050--8057, 2019.

\bibitem[Ranganath et~al.(2013)Ranganath, Jurafsky, and
  McFarland]{ranganath-jurafsky}
Rajesh Ranganath, Dan Jurafsky, and Daniel~A. McFarland.
\newblock Detecting friendly, flirtatious, awkward, and assertive speech in
  speed-dates.
\newblock \emph{Computer Speech and Language}, 27\penalty0 (1):\penalty0
  89--115, 2013.

\bibitem[Reddy(1979)]{reddy1979conduit}
Michael Reddy.
\newblock The conduit metaphor --- {A} case of frame conflict in our language
  about language.
\newblock \emph{Metaphor and Thought}, 2:\penalty0 285--324, 1979.

\bibitem[Roberts and Francis(2013)]{roberts13}
Felicia Roberts and Alexander~L. Francis.
\newblock Identifying a temporal threshold of tolerance for silent gaps after
  requests.
\newblock \emph{Journal of the Acoustical Society of America}, 133:\penalty0
  471--477, 2013.

\bibitem[Sarikaya(2017)]{sarikaya2017overview}
Ruhi Sarikaya.
\newblock The technology behind personal digital assistants: An overview of the
  system architecture and key components.
\newblock \emph{IEEE Signal Processing Magazine}, 34\penalty0 (1):\penalty0
  67--81, 2017.

\bibitem[Schlangen(2004)]{schlangen2004causes}
David Schlangen.
\newblock Causes and strategies for requesting clarification in dialogue.
\newblock In \emph{Proceedings of SIGdial}, pages 136--143, 2004.

\bibitem[Schlangen and Skantze(2011)]{Schlangen2011}
David Schlangen and Gabriel Skantze.
\newblock {A General, Abstract Model of Incremental Dialogue Processing}.
\newblock \emph{Dialogue {\&} Discourse}, 2\penalty0 (1):\penalty0 83--111,
  2011.

\bibitem[Sheridan(2016)]{sheridan2016}
Thomas~B. Sheridan.
\newblock Human-robot interaction: Status and challenges.
\newblock \emph{Human Factors}, 58\penalty0 (4):\penalty0 525--532, 2016.

\bibitem[Skantze(2005)]{skantze2005exploring}
Gabriel Skantze.
\newblock Exploring human error recovery strategies: Implications for spoken
  dialogue systems.
\newblock \emph{Speech Communication}, 45\penalty0 (3):\penalty0 325--341,
  2005.

\bibitem[Skantze et~al.(2015)Skantze, Johansson, and
  Beskow]{skantze2015exploring}
Gabriel Skantze, Martin Johansson, and Jonas Beskow.
\newblock Exploring turn-taking cues in multi-party human-robot discussions
  about objects.
\newblock In \emph{Proceedings of ICMI}, pages 67--74, 2015.

\bibitem[Smith et~al.(2015)Smith, Chao, and Thomaz]{smith-chao}
Justin~S. Smith, Crystal Chao, and Andrea~L. Thomaz.
\newblock Real-time changes to social dynamics in human-robot turn-taking.
\newblock In \emph{Proceedings of IROS}, pages 3024--3029, 2015.

\bibitem[Tan et~al.(2019)Tan, Yu, and Bansal]{tan-etal-2019-learning}
Hao Tan, Licheng Yu, and Mohit Bansal.
\newblock Learning to navigate unseen environments: Back translation with
  environmental dropout.
\newblock In \emph{Proceedings of NAACL}, pages 2610--2621, 2019.

\bibitem[Tannen(1989)]{tannen-meant}
Deborah Tannen.
\newblock \emph{That's Not What I Meant! How Conversational Style Makes or
  Breaks Relationships}.
\newblock Ballantine, 1989.

\bibitem[Tellex et~al.(2011{\natexlab{a}})Tellex, Kollar, Dickerson, Walter,
  Banerjee, Teller, and Roy]{tellex2011approaching}
Stefanie Tellex, Thomas Kollar, Steven Dickerson, Matthew~R. Walter,
  Ashis~Gopal Banerjee, Seth Teller, and Nicholas Roy.
\newblock Approaching the symbol grounding problem with probabilistic graphical
  models.
\newblock \emph{AI Magazine}, 32\penalty0 (4):\penalty0 64--76,
  2011{\natexlab{a}}.

\bibitem[Tellex et~al.(2011{\natexlab{b}})Tellex, Kollar, Dickerson, Walter,
  Banerjee, Teller, and Roy]{tellex2011understanding}
Stefanie Tellex, Thomas Kollar, Steven Dickerson, Matthew~R. Walter,
  Ashis~Gopal Banerjee, Seth Teller, and Nicholas Roy.
\newblock Understanding natural language commands for robotic navigation and
  mobile manipulation.
\newblock In \emph{Proceedings of AAAI}, pages 1507--1514, 2011{\natexlab{b}}.

\bibitem[Tellex et~al.(2020)Tellex, Gopalan, Kress-Gazit, and
  Matuszek]{tellex2020robots}
Stefanie Tellex, Nakul Gopalan, Hadas Kress-Gazit, and Cynthia Matuszek.
\newblock Robots that use language.
\newblock \emph{Annual Review of Control, Robotics, and Autonomous Systems},
  3:\penalty0 25--55, 2020.

\bibitem[Thomason et~al.(2016)Thomason, Sinapov, Svetlik, Stone, and
  Mooney]{thomason2016learning}
Jesse Thomason, Jivko Sinapov, Maxwell Svetlik, Peter Stone, and Raymond~J.
  Mooney.
\newblock Learning multi-modal grounded linguistic semantics by playing ``{I
  Spy}''.
\newblock In \emph{Proceedings of IJCAI}, pages 3477--3483, 2016.

\bibitem[Thomason et~al.(2020)Thomason, Padmakumar, Sinapov, Walker, Jiang,
  Yedidsion, Hart, Stone, and Mooney]{thomason:jair20}
Jesse Thomason, Aishwarya Padmakumar, Jivko Sinapov, Nick Walker, Yuqian Jiang,
  Harel Yedidsion, Justin Hart, Peter Stone, and Raymond~J. Mooney.
\newblock Jointly improving parsing and perception for natural language
  commands through human-robot dialog.
\newblock \emph{Journal of Artificial Intelligence Research}, 67:\penalty0
  327--374, 2020.

\bibitem[Vinciarelli et~al.(2012)Vinciarelli, Pantic, Heylen, Pelachaud, Poggi,
  D'Errico, and Schroeder]{vinciarelli12}
Alessandro Vinciarelli, Maja Pantic, Dirk Heylen, Catherine Pelachaud, Isabella
  Poggi, Francesca D'Errico, and Marc Schroeder.
\newblock Bridging the gap between social animal and unsocial machine: A survey
  of social signal processing.
\newblock \emph{IEEE Transactions on Affective Computing}, 3\penalty0
  (1):\penalty0 69--87, 2012.

\bibitem[Wang et~al.(2018)Wang, Stanton, Zhang, Skerry-Ryan, Battenberg, Shor,
  Xiao, Ren, Jia, and Saurous]{style-tokens}
Yuxuan Wang, Daisy Stanton, Yu~Zhang, RJ~Skerry-Ryan, Eric Battenberg, Joel
  Shor, Ying Xiao, Fei Ren, Ye~Jia, and Rif~A. Saurous.
\newblock Style tokens: Unsupervised style modeling, control and transfer in
  end-to-end speech synthesis.
\newblock In \emph{Proceedings of ICML}, pages 5180--5189, 2018.

\bibitem[Ward(2019)]{ward-book}
Nigel~G. Ward.
\newblock \emph{Prosodic Pattterns in English Conversation}.
\newblock Cambridge University Press, 2019.

\bibitem[Ward and DeVault(2016)]{ward-devault-aimag}
Nigel~G. Ward and David DeVault.
\newblock Challenges in building highly-interactive dialog systems.
\newblock \emph{AI Magazine}, 37\penalty0 (4):\penalty0 7--18, 2016.

\bibitem[Wilson and Moore(2017)]{Wilson2017}
Sarah Wilson and Roger~K. Moore.
\newblock Robot, alien and cartoon voices: Implications for speech-enabled
  systems.
\newblock In \emph{Proceedings of the 1st International Workshop on Vocal
  Interactivity in-and-between Humans, Animals and Robots (VIHAR)}, pages
  40--44, 2017.

\bibitem[Wiltshire et~al.(2013)Wiltshire, Barber, and
  Fiore]{wiltshire2013towards}
Travis~J. Wiltshire, Daniel Barber, and Stephen~M. Fiore.
\newblock Towards modeling social-cognitive mechanisms in robots to facilitate
  human-robot teaming.
\newblock In \emph{Proceedings of HFES}, volume~57, pages 1278--1282, 2013.

\bibitem[Wolfgang et~al.(2017)Wolfgang, Lukic, Sander, Martin, and
  K\"{u}pper]{Wolfgang2017}
Meldon Wolfgang, Vladimir Lukic, Alison Sander, Joe Martin, and Daniel
  K\"{u}pper.
\newblock Gaining robotics advantage.
\newblock The Boston Consulting Group (BCG), 2017.

\bibitem[Yang et~al.(2018)Yang, Bellingham, Dupont, Fischer, Floridi, Full,
  Jacobstein, Kumar, McNutt, Merrifield, Nelson, Scassellati, Taddeo, Taylor,
  Veloso, Wang, and Wood]{yang2018grand}
Guang-Zhong Yang, Jim Bellingham, Pierre~E. Dupont, Peer Fischer, Luciano
  Floridi, Robert Full, Neil Jacobstein, Vijay Kumar, Marcia McNutt, Robert
  Merrifield, Bradley~J. Nelson, Brian Scassellati, Mariarosaria Taddeo,
  Russell Taylor, Manuela Veloso, Zhong~Lin Wang, and Robert Wood.
\newblock The grand challenges of science robotics.
\newblock \emph{Science Robotics}, 3\penalty0 (14), 2018.

\bibitem[Yu et~al.(2015)Yu, Bohus, and Horvitz]{yu-bohus15}
Zhou Yu, Dan Bohus, and Eric Horvitz.
\newblock Incremental coordination: Attention-centric speech production in a
  physically situated conversational agent.
\newblock In \emph{Proceedings of SIGdial}, pages 402--406, 2015.

\end{thebibliography}
